%% file: main.tex
\PassOptionsToPackage{table}{xcolor}
\documentclass[sigconf, 9pt]{acmart}

\AtBeginDocument{%
  }

\copyrightyear{2025}
\acmYear{2025}
\setcopyright{acmlicensed}
\acmConference[WWW Companion '25]{Companion Proceedings of the ACM Web Conference 2025}{April 28-May 2, 2025}{Sydney, NSW, Australia}
\acmBooktitle{Companion Proceedings of the ACM Web Conference 2025 (WWW Companion '25), April 28-May 2, 2025, Sydney, NSW, Australia}
\acmISBN{979-8-4007-1331-6/25/04}
\acmDOI{10.1145/3701716.3715245}

\usepackage{algorithm}
\usepackage{algorithmic}
\usepackage{array}
\usepackage{enumitem}
\usepackage{graphics}
\usepackage{graphicx}
\usepackage{makecell}
\usepackage{multirow}
\usepackage{subfigure}

\usepackage{titlesec}
\titleformat{\section}{\normalfont\Large\bfseries}{\thesection}{1em}{}

\begin{document}

\title{Few-shot LLM Synthetic Data with Distribution Matching}




\author{Jiyuan Ren}
\email{rjy22@mails.tsinghua.edu.cn}
\affiliation{%
  \institution{Tsinghua University}
  \city{Beijing}
  \country{China}}
\authornote{Both authors contributed equally to this research.}

\author{Zhaocheng Du}
\email{zhaochengdu@huawei.com}
\affiliation{%
  \institution{Huawei Noah's Ark Lab}
  \city{Shenzhen}
  \country{China}}
\authornotemark[1]

\author{Zhihao Wen}
\email{wenzhihao4@huawei.com}
\affiliation{%
  \institution{Huawei Noah's Ark Lab}
  \city{Singapore}
  \country{Singapore}}

\author{Qinglin Jia}
\email{jiaqinglin2@huawei.com}
\affiliation{%
  \institution{Huawei Noah's Ark Lab}
  \city{Beijing}
  \country{China}}

\author{Sunhao Dai}
\email{sunhaodai@ruc.edu.cn}
\affiliation{%
  \institution{Renmin University of China}
  \city{Beijing}
  \country{China}}

\author{Chuhan Wu}
\email{wuchuhan@huawei.com}
\affiliation{%
  \institution{Huawei Noah's Ark Lab}
  \city{Beijing}
  \country{China}}

\author{Zhenhua Dong}
\email{dongzhenhua@huawei.com}
\affiliation{%
  \institution{Huawei Noah's Ark Lab}
  \city{Shenzhen}
  \country{China}}
\authornote{Corresponding author.}
\renewcommand{\shortauthors}{Ren and Du et al.}

\begin{abstract}
As large language models (LLMs) advance, their ability to perform in-context learning and few-shot language generation has improved significantly. This has spurred using LLMs to produce high-quality synthetic data to enhance the performance of smaller models like online retrievers or weak LLMs. 
However, LLM-generated synthetic data often differs from the real data in key language attributes (e.g., styles, tones, content proportions, etc.). 
As a result, mixing these synthetic data directly with real data may distort the original data distribution, potentially hindering performance improvements. 
To solve this, we introduce \textbf{SynAlign}: a synthetic data generation and filtering framework based on key attribute distribution matching.
Before generation, SynAlign employs an uncertainty tracker surrogated by the Gaussian Process model to iteratively select data clusters distinct from selected ones as demonstrations for new data synthesis, facilitating the efficient exploration diversity of the real data. Then, a latent attribute reasoning method is employed: the LLM summarizes linguistic attributes of demonstrations and then synthesizes new data based on them. This approach facilitates synthesizing diverse data with linguistic attributes that appear in real data.
After generation, the Maximum Mean Discrepancy is used as the objective function to learn the sampling weight of each synthetic data, ensuring distribution matching with the real data. Our experiments on multiple text prediction tasks show significant performance improvements. We also conducted an online A/B test on an online retriever to demonstrate SynAlign's effectiveness. Our code is available \href{https://github.com/nighood/SynAlign}{\textcolor{blue}{here}}.
\end{abstract}




\maketitle

\input{sections/1_introduction}
\input{sections/2_relatedwork}
\input{sections/3_method}
\input{sections/4_experiments}
\input{sections/5_conclusion}

\bibliographystyle{ACM-Reference-Format}
\balance
\bibliography{ref}

\clearpage
\appendix

\input{sections/6_appendix}

\end{document}

%% file: sections/1_introduction.tex
\section{Introduction}
Despite the rapid development of large language models (LLMs), there remains a demand in the industry for smaller online language models tailored to high-latency scenarios~\cite{hsieh2023distilling, jia2024erase, wang2023single, du2024lightcs, du2024tutorial, zhao2024retrievable, dai2023uncovering, dai2024modeling} like search engines~\cite{khattab2020colbert}. These models are typically trained on pre-constructed datasets for online services. 
However, real-world data collection often suffers from various biases like selection bias~\cite{wang2020exposure, arora2022exposure, dai2023dually} and long-tail issues~\cite{dai2023long, wu2008information}, and expensive manual cost~\cite{daniel2018quality, dai2024cocktail}. Synthetic data generation has the potential to mitigate these biases~\cite{shahul2024bias, lyu2022semi}, enhance data diversity~\cite{feng2020genaug}, and ultimately improve model generalization and accuracy.

To fully harness the potential of synthetic data, it is essential to define how synthetic data can be high quality for a specific task? High quality can be assessed along various dimensions, such as low noise and fairness. However, in profit-driven industrial applications like search engines, we prioritize improving model accuracy as the objective. Based on Murphy’s research \cite{murphy2012machine}, we propose the following definition of high quality: \textbf{the optimal dataset is that which most closely matches the distribution under which the model will be evaluated}. Under this definition, synthetic high-quality data requires matching real data in some key attributes.

\begin{figure}[!t]
  \centering
    \includegraphics[width=1\linewidth]{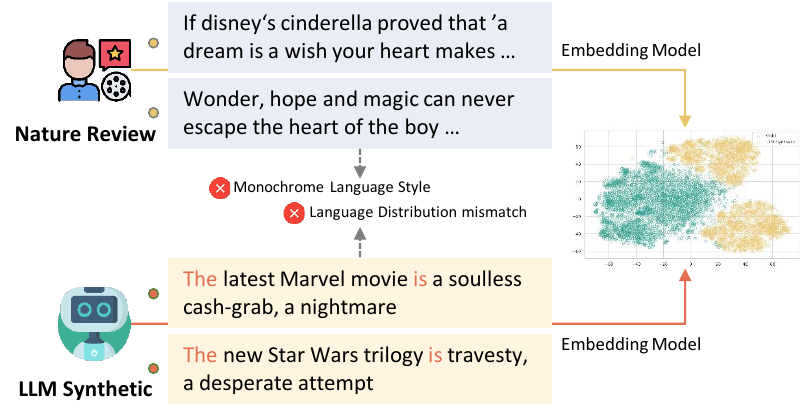}
  \caption{A case on how LLM synthetic samples misalign with human-generated samples.}
  \vspace{-0.15in}
  \label{fig.motivation}
  \Description{}
\end{figure}

Many existing methods adopted distribution matching as the objective function and used all or a subset of the real dataset to train a generative model that approximates the real data distribution, and then synthesizes data based on this model. For instance, RelGAN ~\cite{RelGAN} designs a generative adversarial network architecture for text data generation tasks, ensuring that the generated data closely aligns with the original data distribution. Similarly, FewGen~\cite{fewgen} fine-tunes an autoregressive pre-trained language model on a small sample dataset and employs it as a generator to synthesize a large number of new training samples, thereby augmenting the original training set. However, such methods require training a dedicated generative model for each scenario, incurring high manual costs. Additionally, due to the risk of catastrophic forgetting~\cite{ramasesh2021effect}, these generators may lose their in-context learning capability. 

Benefiting from massive high-quality training data and carefully designed training processes~\cite{ouyang2022training}, LLMs have developed strong instruction following and in-context learning abilities that can generate responses for diverse tasks~\cite{brown2020language}. This has prompted academia and industry to use LLMs for zero-shot data synthetic to train their domain-specific models. For example, GPT-3Mix ~\cite{GPT3Mix} utilizes large-scale language models to generate mixed samples, enhancing text datasets. Additionally, AttrPrompt ~\cite{attriprompt} generates training data through diverse attribute prompts, improving model performance while reducing data bias.

However, experiments (see section \ref{exp:datamismatch}) show two issues with the data synthesized by LLMs from the distribution matching perspective. First, LLM's zero-shot generation results cannot cover all real data's linguistic attributes (especially those ``imperfect'' ones like typos or incomplete sentences). Secondly, there is a significant discrepancy in the proportion of linguistic attributes between the LLMs synthetic data and the real data ~\cite{li2023synthetic, gao2025samplellm}. Both are primarily due to LLM’s limitations in understanding long text inputs, which prevents it from incorporating distributions of real data and previously synthetic data. These issues result in a considerable divergence between LLM-synthetic and real data, meaning that directly adding these synthetic data to the real data may not not improve model performance.

To address these challenges, we designed \textbf{SynAlign} (\textbf{Syn}thetic data generation and distribution \textbf{Align}ment framework), which incorporates three modules during LLMs' synthetic data generation process: Exploration-aware Sampling, Latent Attribute Reasoning, and Synthetic Distribution Alignment. 
Exploration-aware sampling is applied to the demonstration sampling process. It utilizes a Gaussian Process model as the samples' uncertainty-tracker. Samples with the highest uncertainty will be selected as demonstrations and their uncertainty will be updated after generation. With this process, data sampling module can efficiently explore real data distribution.
Selected demonstrations will be fed into the Latent Attribute Reasoning module to summarize key attributions covering language contents, styles, etc. Afterward, new data are synthesized based on these generalized latent attributes.
After the data synthesis process, the Synthetic Distribution Alignment module employs a post-training approach to assign sampling weights for each synthetic sample by minimizing Maximum Mean Discrepancy between synthetic data and real data. After resampling the synthetic data based on sampling weights, we obtain the final synthetic data that more closely aligns with real data distribution.

Extensive experiments demonstrate that our method synthesizes high-quality data more efficiently, requiring fewer tokens. Compared to existing data generation algorithms, the synthetic data generated by SynAlign consistently achieves superior performance. The main contributions of this paper are as follows:

$\bullet$ We designed an Exploration-aware Sampling module that can efficiently explore all language attributes in real data. Data synthesized with latent attributes extracted from these samples could cover all real data's distribution.  

$\bullet$ We designed a Synthetic Distribution Alignment module to post-align synthetic data distribution with real data distribution.

$\bullet$  Extensive experiments are conducted and demonstrate our method can generate high quality synthetic samples efficiently.

%% file: sections/2_relatedwork.tex
\section{Related Work}

The advent of LLMs has revolutionized synthetic data generation, offering significant advantages over traditional methods. LLMs excel in generating coherent and human-like text, making them effective tools for creating high-quality datasets~\citep{survey_llmsdrivensyntheticdata}. Compared to manual data collection, LLMs provide notable benefits in flexibility, efficiency, and scalability. They can tailor datasets to specific needs by adjusting prompts and conditions~\citep{tinystories}, significantly reduce annotation costs~\citep{self_improve}, and automate the training pipeline, enabling broader application across multiple domains~\citep{sungen}.

Despite these advantages, ensuring the quality and relevance of LLM-generated datasets requires robust generation techniques and curation strategies. Below, we review key methods for data generation, focusing on prompt engineering and multi-step generation, followed by strategies for data curation and distribution alignment.

\subsection{LLM-Based Data Generation Methods}

\paragraph{Prompt Engineering.}  
Prompt engineering is critical for controlling the quality and diversity of synthetic data. Effective prompts typically define tasks clearly, specify generation conditions (e.g., themes or styles), and include in-context demonstrations, which help guide LLMs toward accurate outputs~\citep{ChatGPTOC, best_practice}. For instance, \citet{wang2023let} demonstrated how condition-based prompts improve stylistic diversity, while \citet{li2023synthetic} highlighted the role of examples in enhancing task alignment.

\paragraph{Multi-Step Generation.}  
Multi-step generation addresses complex data needs by breaking the process into sub-tasks. Sample-wise decomposition divides data into smaller parts for step-by-step generation, improving coherence, as shown by \citet{he2023annollm}. Dataset-wise decomposition dynamically adjusts generation conditions to enhance diversity and coverage~\citep{wang2023let}. Our method builds on these approaches by introducing systematic condition controls to further improve data quality and diversity.

\subsection{Data Curation and Distribution Alignment}

While LLMs excel at generating diverse data, synthetic datasets often contain noise or distributional biases that can hinder downstream performance~\citep{regen}. To address these issues, curation strategies such as sample filtering and label enhancement have been proposed. Sample filtering uses heuristic metrics like confidence scores or influence functions to identify high-quality samples~\citep{seedat2023curated, progen}. Label enhancement techniques, such as knowledge distillation~\citep{xiao2023freeal}, refine labels\citep{wan2024tnt} and reduce annotation errors\citep{li2023coannotating}. 

To align synthetic data distributions with real-world data, methods like Maximum Mean Discrepancy (MMD) have been employed~\citep{survey_llmsdrivensyntheticdata}. Our approach integrates systematic sample selection and MMD-based alignment, ensuring the synthetic data closely resembles real-world distributions while maintaining high quality.


%% file: sections/3_method.tex
\section{Method}
\label{sec_method}

\begin{figure*}[!t]
  \centering
    \includegraphics[width=0.99\linewidth]{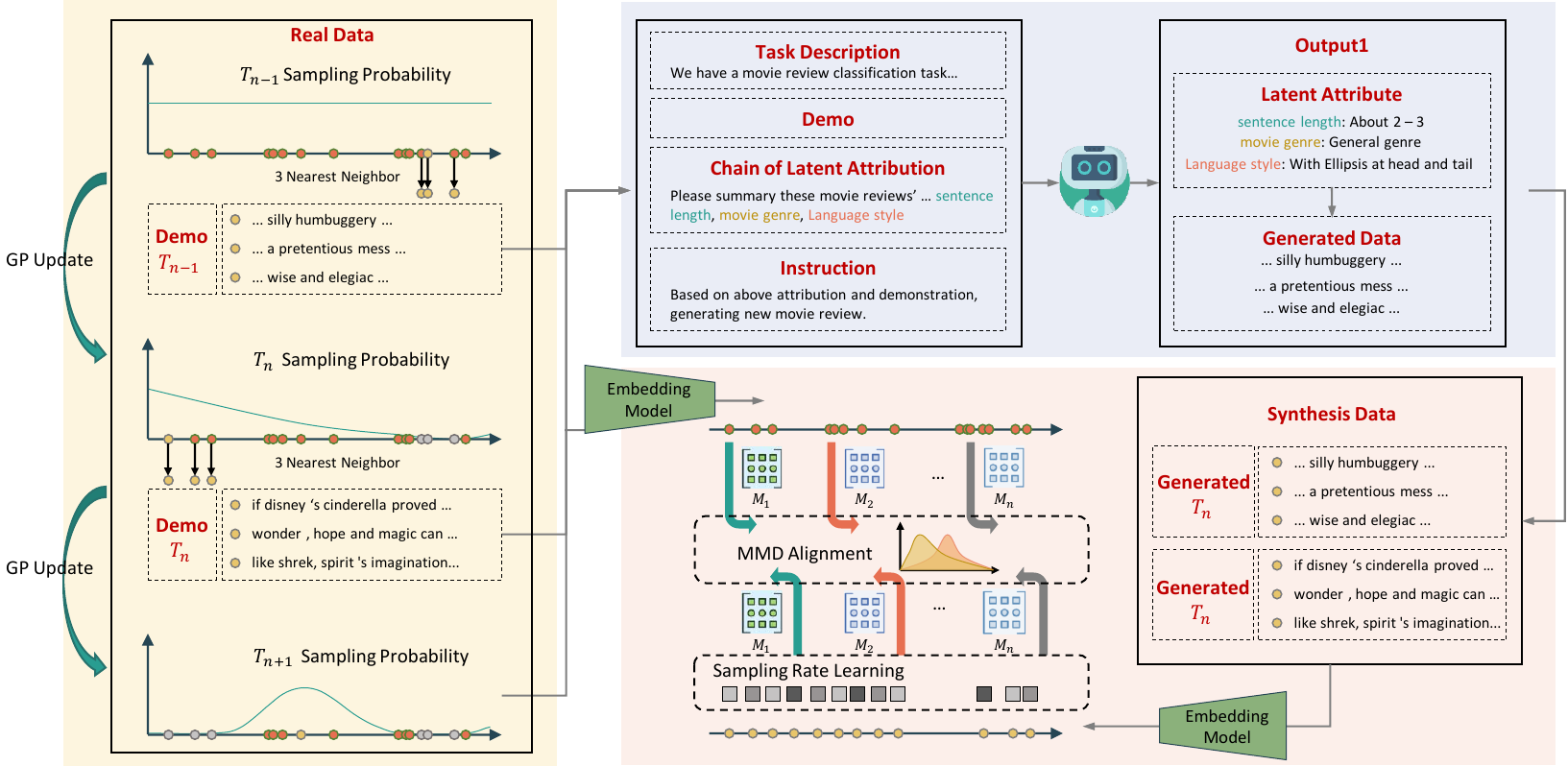}
  \caption{SynAlign comprises three modules for aligning the distribution of synthetic samples. First, the \textbf{Exploration-aware Sampling} Module selects real samples based on uncertainty to provide diverse inputs for the LLM. Next, the \textbf{Latent Attribute Reasoning} Module uses these samples as demonstrations to identify and generalize key language attributes for synthetic data generation. Finally, the \textbf{Synthetic Distribution Alignment} Module assigns sampling weights to synthetic data, which are then resampled accordingly to ensure alignment with the real data distribution.}
  \label{fig.model}
  \Description{}
\end{figure*}

\subsection{Overview}

Given a real dataset $D_{\text{ori}} = \{(x_i, y_i)\}_{i=1}^N$ and a synthetic dataset $D_{\text{gen}} = \{(x'_j, y'_j)\}_{j=1}^M$ generated by a large language model (LLM), our goal is to enhance the quality and diversity of $D_{\text{gen}}$ such that it improves the performance of smaller, domain-specific models in downstream tasks. To simplify the modeling of text distributions, both $D_{\text{ori}}$ and $D_{\text{gen}}$ are mapped into embedding spaces $E_{\text{ori}}$ and $E_{\text{gen}}$ using Sentence-BERT~\cite{reimers2019sentence}.

Directly generating $D_{\text{gen}}$ often leads to a distributional gap between real and synthetic data due to LLM limitations. This gap arises from incomplete coverage of real data diversity or misaligned proportions of data attributes, which can degrade the utility of $D_{\text{gen}}$ in downstream tasks.

To address these issues, we propose \textbf{SynAlign}, comprising three key modules:
1) \textbf{Exploration-aware Sampling}: This module uses a Gaussian Process (GP) uncertainty tracker to actively select diverse and representative real samples as demonstrations for synthetic data generation.
2) \textbf{Latent Attribute Reasoning}: Demonstrations are used to reason about key linguistic attributes via a Chain-of-Thought (CoT)~\cite{wei2022chain} paradigm, guiding the LLM to explicitly attend to diverse attributes during generation.
3) \textbf{Synthetic Distribution Alignment}: Using Maximum Mean Discrepancy (MMD)~\cite{gretton2012kernel}, this module aligns $D_{\text{gen}}$ with $D_{\text{ori}}$ by computing sampling weights for synthetic samples, ensuring minimal distributional deviation.

By combining these modules, SynAlign produces high-quality, diverse synthetic data that better aligns with real data distributions, resulting in improved performance in downstream tasks.

\subsection{Exploration-aware Sampling}

Our method uses a few-shot prompting approach to generate synthetic data using LLM. Selecting representative demonstrations from the real dataset is crucial for guiding the LLM in producing high-quality synthetic data. Random selection may overfit to frequently occurring patterns. We utilize an uncertainty-aware sampling strategy to ensure that the synthetic data process efficiently covers all real data's language attributes.


To achieve this, we use a Gaussian Process model ~\cite{williams1995gaussian} as samples' uncertainty tracker $U(E_{ori})$. When initializing $U(E_{ori})$, we assign each sample's mean value as 0 and uncertainty (variance) as 1 and covariance between samples $e_i, e_j$ as $\kappa(e_i, e_j)$.
\begin{align}
                    \left[\begin{array}{c}
                    U\left(e_0\right) \\
                    \vdots \\
                    U\left(e_n\right)
                    \end{array}\right] \sim N\left(\left[\begin{array}{c}
                    0 \\
                    \vdots \\
                    0
                    \end{array}\right],\left[\begin{array}{ccc}
                    1 & \cdots & \kappa\left(e_1, e_n\right) \\
                    \vdots & \cdots & \vdots \\
                    \kappa\left(e_n, e_1\right) & \cdots & 1
                    \end{array}\right]\right) 
\end{align}

Where $\kappa(\cdot, \cdot)$ is chosen as Radial Basis Function Kernel as below to ensure covariance between selected and unselected samples increase as text embedding's similarity decreases.
\begin{align}
\kappa(e_i, e_j) = exp(-\frac{1}{2\tau}||e_i- e_j||)
\end{align} 
Where $\tau$ is a hyper-parameter called bandwidth that controls covariance smoothness. Each demonstration selection phase includes the following two steps:

\textbf{Step 1. Demonstration Selection}. Samples with the highest uncertainty (variance) and its $k$-nearest samples $D_{dem}$ will be selected out as demonstrations according to the newest updated $U(E_{ori})$. These samples are most different to LLM already seen demonstrations regarding linguistic attributes like styles, contexts etc.
\begin{align}
D_{dem} = k\text{-}NN(D_{ori}, argmax_{i}(U(e_i|e_i \in E_{ori})), k)
\end{align} 
Where $k\text{-}NN$ is the function that returns the $k$ nearest samples of the most uncertain sample indexed by $argmax_{i}(U(e_i|e_i \in E_{ori}))$ from real dataset $D_{ori}$.

\textbf{Step 2. Uncertainty Update}. Once $D_{dem}$ are used as demonstrations for data synthetic, their embedding $E_{dem}$'s uncertainty is reduced to 0 and constructed as posterior training data $(E_{dem}, 0)$. Those samples together with historical ones $(E_{s}, 0)$ will be used to update the unselected samples' uncertainty value $U(E_{u})$ in the uncertainty tracker $U$. The updated sample uncertainty is given below:
\begin{align}
U(E_{u}) | E_{u}, E_{s}, U(E_{s}) \sim N\left(\mu^*,\Sigma^*\right) 
\end{align}                          
where $\mu^*$ equals to \textbf{0} because only uncertainty is used in the selection process. And variance(uncertainty) $\Sigma^*$ is given below:
\begin{align}
\Sigma^* = K(E_{u}, E_{u}) + I -K(E_{u}, E_{s})K(E_{s}, E_{s} + I)^{-1}K(E_{s}, E_{u}) 
\end{align} 
These two steps are repeated iteratively to select demonstrations for LLM to generate synthetic data until all samples' uncertainties are below a predefined threshold. The overall procedure is listed in Appendix Algorithm~\ref{apdx:alg}.

This uncertainty-aware sampling strategy ensures that the selected demonstrations represent the diversity of real data distribution, as measured by the uncertainty in the Gaussian Process model. These selected examples are then used as demonstrations for the few-shot prompting of the LLM in the next stage of the data generation process.

\subsection{Latent Attribute Reasoning}
Demonstrations selected from the previous module are used to feed and assist LLM in understanding the diverse linguistic attributes in real data so that the synthetic data produced by LLM won't be monochrome. To explicitly ensure the synthetic data captures all linguistic attributes in the real data while maintaining content diversity, a two-stage process was designed by first reasoning out linguistic attributes and then generating diverse content based on them. 

\textbf{Stage 1. Key Attribute Reasoning} stage. The goal of this stage is to identify and summarize the key linguistic attributes of the selected demonstrations, which serve as a blueprint for synthetic data generation. To ensure the synthetic data $D_{gen}$ reflects the structure and diversity of the real data $D_{ori}$, we let $D_{gen}$ mimick key attributes $\textbf{A} = \{a_1,..a_n\}$ extracted from $D_{dem}$. Referring to prior work AttrPrompt ~\cite{attriprompt}, we use LLM to identify crucial attributes. For instance, attributes in an Amazon product review dataset might include \textit{Product Info, Usage Experience,} and \textit{Writing Style}. These attributes provide a framework for summarizing the sampled examples.

Once the key attributes are identified, we construct reasoning prompts $P_1$ instructing the LLM to analyze $D_{dem}$ and extract these key attributes. This results in a JSON format \textit{Attribute Summary Set} $S$:
\begin{align}
S = LLM(D_{dem}, A, P_1) =\{(a_1,v_1), (a_2,v_2)... (a_n,v_n)\}
\end{align} 
where each tuple $(a_i,v_i)$ represents summarized attributes of selected demonstrations.

\textbf{Stage 2. Attribute-Based Data Generation} stage. We leverage the attribute summarized in Stage 1 to guide the LLM in generating synthetic data. By incorporating attribute summaries $S$ into generation prompts $P_2$, LLM can synthesize samples reflecting these key attributes, such as product information and writing style. For each attribute tuple in $\mathcal{S}$, the LLM generates a synthetic sample that adheres to the specified attributes, ensuring the generated data is both diverse and representative of the original dataset. Finally, by generating new data for each attribute summary, we construct the synthetic dataset $D_{\text{gen}}$.
\begin{align}
D_{\text{gen}} = LLM(S,P_2) =\{(x_0, y_0), (x_1, y_1), \dots, (x_M, y_M)\}
\end{align}
 The size of $D_{\text{gen}}$ depends on the number of summaries in $\mathcal{S}$ and the samples generated per summary. The generated data covers a wide range of latent attributes, maintaining alignment with the original data distribution while introducing new variations in style and content.

\subsection{Synthetic Distribution Alignment}

Due to the limited input length, LLMs cannot fully account for the distribution of $D_{gen}$ and $D_{ori}$. This limitation often results in discrepancies between the linguistic attribute distributions of the synthesized and real data, potentially causing a "seesaw effect" that degrades the model's accuracy in practical applications. To enhance distribution matching, we want to learn a post-transformational function $F_\omega(\cdot)$ to minimize the distance between these two distributions:
\begin{align}
argmin_{F_\omega(\cdot)}\ Dist(D_{\text{ori}} || F_\omega(D_{\text{gen}}))
\end{align}
However, Due to the discrete and high-dimensional nature of linguistic data, measuring their distribution exactly is intractable. To address this, we adopted the Maximum Mean Discrepancy (MMD) method to approximate this matching objective. The basic idea of MMD in our application is matching the mean embedding of $E_{gen}$ and $E_{ori}$ projected in Reproducing Kernel Hilbert Space (RKHS), which is equivalent to matching these two distributions ~\cite{zhao2023dataset}.  
\begin{align}
argmin_{F_\omega(\cdot)}
sup_{||\phi_\theta||_H\leq 1} (
E\left [\phi_\theta(D_{\text{ori}})\right ] - 
E\left [\phi_\theta (F_\omega(D_{\text{gen}})\right ])
\end{align}
Where the $\phi_\theta$ is a family of functions parameterized by $\theta$ and $H$ represent the RKHS, considering the ground truth distribution is intangible, we adopt its empirical approximation by making the following changes: (1) map the text into an embedding space to obtain continuous representations; (2) simplify the transformation function as a data sampling weight $\omega$ and (3) assuming $\phi_\theta(\cdot)$ as a $R^n\rightarrow R^1$ random linear projection matrix family $\Theta$. Finally, we can derive the following objective function.
 \begin{align}
argmin_{\omega} E_{\theta\sim \Theta}||\frac{1}{N}\sum_{i=1}^{N}\theta\cdot E_{\text{ori}} - \frac{1}{M}\sum_{i=1}^{M}\theta\cdot (\omega \cdot E_{\text{gen}}))||^2
\label{equationset}
\end{align}
To ensure diversity in the projections matrix set $\Theta$, we use Gram-Schmidt orthogonalization\citep{Gram_Schmidt} to initialize these random matrices. The parameters of each projection matrix $\theta_{i}$ are orthogonalized regarding the previously initialized matrix $\theta_{1:i-1}$, ensuring each network captures different aspects of the data.
 \begin{align}
\Theta = \{\theta_i|<\theta_i,\theta_j>=\delta_{ij}\forall i, j, \delta_{ij} =1\ if\ i==j,\ otherwise\ 0\}
\end{align}
The final sample weight of each synthetic sample $\omega$ is calculated by solving the linear equation set described in formula ~\ref{equationset}. In our implementation, we use gradient descent to solve $\omega$ iteratively.

After the importance weight of each sample $\omega$ has been learned in the distribution alignment process, we can perform re-sampling on the original synthetic dataset $D_{gen}$ based on $\omega$ with replacement to construct a new synthetic dataset $D'_{gen}$ which will have similar linguistic attribute distribution with the data $D_{ori}$. Mixing up the $D'_{gen}$ with the original data $D_{ori}$ can bring higher model performance than using $D_{gen}$. The pseudocode of our algorithm is given in \ref{apdx:alg}

%% file: sections/4_experiments.tex
\section{Experiments}
\subsection{Experimental Setup}


We evaluate our method on three widely-used text classification datasets: SST-2~\cite{sst2}, AGNEWS~\cite{agnews}, and Amazon~\cite{amazon}, covering diverse tasks such as sentiment analysis, topic classification, and product review classification. These datasets are chosen for their varying challenges, including class imbalance and linguistic diversity, making them ideal for testing synthetic data generation methods. Table~\ref{tab:data_info} summarizes their key characteristics.

To assess model performance, we adopt two standard metrics: Accuracy (Acc), which measures the proportion of correctly classified samples, and the F1 Score, calculated as the macro-average across all classes. The latter is particularly useful for datasets with imbalanced class distributions. 

We compare it against several baselines, as summarized below.
\begin{itemize}
    \item \textbf{Gold}: Models are trained solely on the original dataset without any synthetic data. 
    \item \textbf{SimPrompt}~\cite{chen2023mixturesoftpromptscontrollable}: Augments the original dataset with $\zeta$ synthetic samples generated using simple class-conditional prompts.
    \item \textbf{AttrPrompt}~\cite{attriprompt}: Uses attribute-rich prompts to generate diverse synthetic data.
    \item \textbf{SynAlign(all)}: Trains models on the full synthetic dataset $D_{\text{gen}}$ without distribution alignment. 
    \item \textbf{SynAlign(random)}: Trains models on the original dataset combined with $\zeta$ randomly selected samples from $D_{\text{gen}}$. 
    \item \textbf{SynAlign(mmd)}: Our proposed method, which selects $\zeta$ samples from $D_{\text{gen}}$ using the MMD-based sampling approach described in Section 3.
\end{itemize}



We fine-tune pre-trained models, including DistilBERT ~\cite{sanh2019distilbert} and BERT-base-uncased ~\cite{bert}, to evaluate the generalization of our method across different model scales. The synthetic data $D_{\text{gen}}$ is generated following the pipeline described in Section \ref{sec_method}. Details on implementation, including training hyperparameters and optimization, are provided in the \ref{apdx:imp_detail}.

\subsection{Main Experimental Results}

\begin{table*}[t]
\centering
\caption{Main results across three datasets. We report Accuracy (Acc) and F1 score for each method and model. The best results are highlighted in bold.}
\label{tab:main_result}
\resizebox{0.8\textwidth}{!}{
\begin{tabular}{llcccccc}
    \toprule
    & & \multicolumn{2}{c}{\textbf{SST-2}} & \multicolumn{2}{c}{\textbf{AGNews}} & \multicolumn{2}{c}{\textbf{Amazon}} \\
    \textbf{Method} & \textbf{Model} & \textbf{Acc} & \textbf{F1} & \textbf{Acc} & \textbf{F1} & \textbf{Acc} & \textbf{F1} \\
    \midrule
    \textbf{LLM zero-shot} &  & 0.9351 & 0.9335 & 0.8232 & 0.8264 & 0.7556 & 0.732 \\
    \midrule\midrule
    \textbf{Gold} & BERT-base-uncased & 0.9248 & 0.9246 & 0.9286 & 0.9398 & 0.8204 & 0.7961 \\
     & DistilBERT & 0.9044 & 0.9045 & 0.9260 & 0.9248 & 0.8053 & 0.8053 \\
     \hline
    \textbf{SimPrompt} & BERT-base-uncased & 0.9264 & 0.9259 & 0.9403 & 0.9402 & 0.8274 & 0.8113 \\
     & DistilBERT & 0.9148 & 0.9164 & 0.9346 & 0.9339 & 0.8165 & 0.8165 \\
     \hline
    \textbf{AttrPrompt} & BERT-base-uncased & 0.9260 & 0.9264 & 0.9441 & 0.9417 & 0.8305 & 0.8142 \\
     & DistilBERT & 0.9231 & 0.9215 & 0.9395 & 0.9395 & 0.8262 & 0.8262 \\
     \hline
    \textbf{SynAlign(all)} & BERT-base-uncased & 0.9292 & 0.9299 & 0.9460 & 0.9441 & 0.8374 & 0.8244 \\
     & DistilBERT & 0.9252 & 0.9212 & 0.9456 & 0.9451 & 0.8351 & 0.8351 \\
     \hline
    \textbf{SynAlign(random)} & BERT-base-uncased & 0.9286 & 0.9322 & 0.9429 & 0.9455 & 0.8204 & 0.8031 \\
     & DistilBERT & 0.9203 & 0.9207 & 0.9441 & 0.9442 & 0.8296 & 0.8296 \\
     \hline
    \textbf{SynAlign(mmd)} & BERT-base-uncased & \textbf{0.9330} & \textbf{0.9352} & \textbf{0.9475} & \textbf{0.9470} & \textbf{0.8381} & \textbf{0.8266} \\
     & DistilBERT & \textbf{0.9282} & \textbf{0.9211} & \textbf{0.9464} & \textbf{0.9433} & \textbf{0.8312} & \textbf{0.8312} \\
    \bottomrule
\end{tabular}}
\end{table*}

Table~\ref{tab:main_result} reports results in terms of \textbf{Accuracy (Acc)} and \textbf{F1 score} for both BERT-base-uncased\cite{bert} and DistilBERT\cite{distilbert}.
The results show that augmenting the original dataset with synthetic data consistently improves performance over the \textbf{Gold} baseline.
This trend is consistent across datasets and models, demonstrating the benefit of synthetic data augmentation. Importantly, \textbf{DistilBERT}, despite its smaller size, achieves competitive results compared to BERT-base-uncased, highlighting the scalability of our method to lightweight models.

Among baseline methods, \textbf{AttrPrompt} generally outperforms \textbf{SimPrompt}, likely due to its ability to generate more diverse and representative synthetic samples. However, both are consistently surpassed by our proposed \textbf{SynAlign} approach, which uses exploration-aware sampling and distribution alignment to select high-quality synthetic data. Within SynAlign, the MMD-based sampling strategy (\textbf{SynAlign(mmd)}) delivers the best results, outperforming \textbf{SynAlign(all)} (which uses all generated samples) and \textbf{SynAlign(random)} (which selects samples randomly). 
These results highlight the importance of informed sample selection, as not all synthetic data contributes equally to performance.

\textbf{SynAlign(mmd)} consistently achieves the highest Accuracy and F1 scores across datasets and models. For example, on SST-2, SynAlign(mmd) with BERT-base-uncased achieves 93.30\% Accuracy, outperforming the Gold baseline (92.48\%) and all other augmentation methods. Similarly, on AGNEWS and Amazon, SynAlign(mmd) delivers superior results, demonstrating robustness across different tasks and domains.

Overall, the experimental results confirm the effectiveness of \textbf{SynAlign(mmd)} in leveraging synthetic data for model improvement. By selectively augmenting datasets with well-aligned samples, our method achieves consistent performance gains across datasets, domains, and model architectures, while maintaining scalability to lightweight models like DistilBERT.


\subsection{Generated Data Analysis}
\label{exp:datamismatch}


\noindent \textbf{Distributional Alignment:}  
We measure the alignment between the original and generated data using \textbf{Wasserstein distance}, which quantifies the cost of transforming one distribution into another. Lower values indicate better alignment. Sentence embeddings are extracted using a pre-trained BERT model and visualized with t-SNE for qualitative analysis.

\begin{figure}[tbp]
    \centering
    \subfigure{\includegraphics[width=0.23\textwidth]{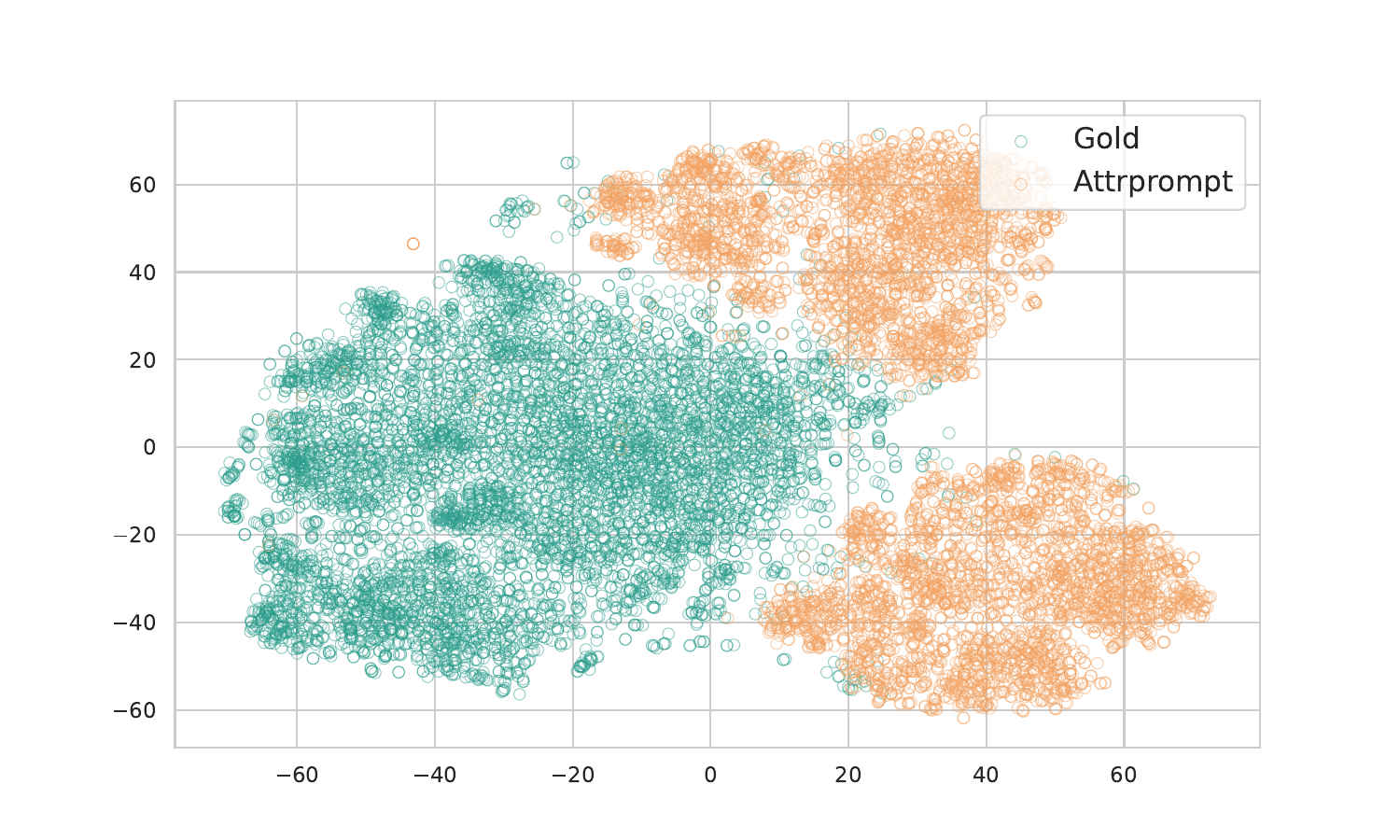}} 
    \subfigure{\includegraphics[width=0.23\textwidth]{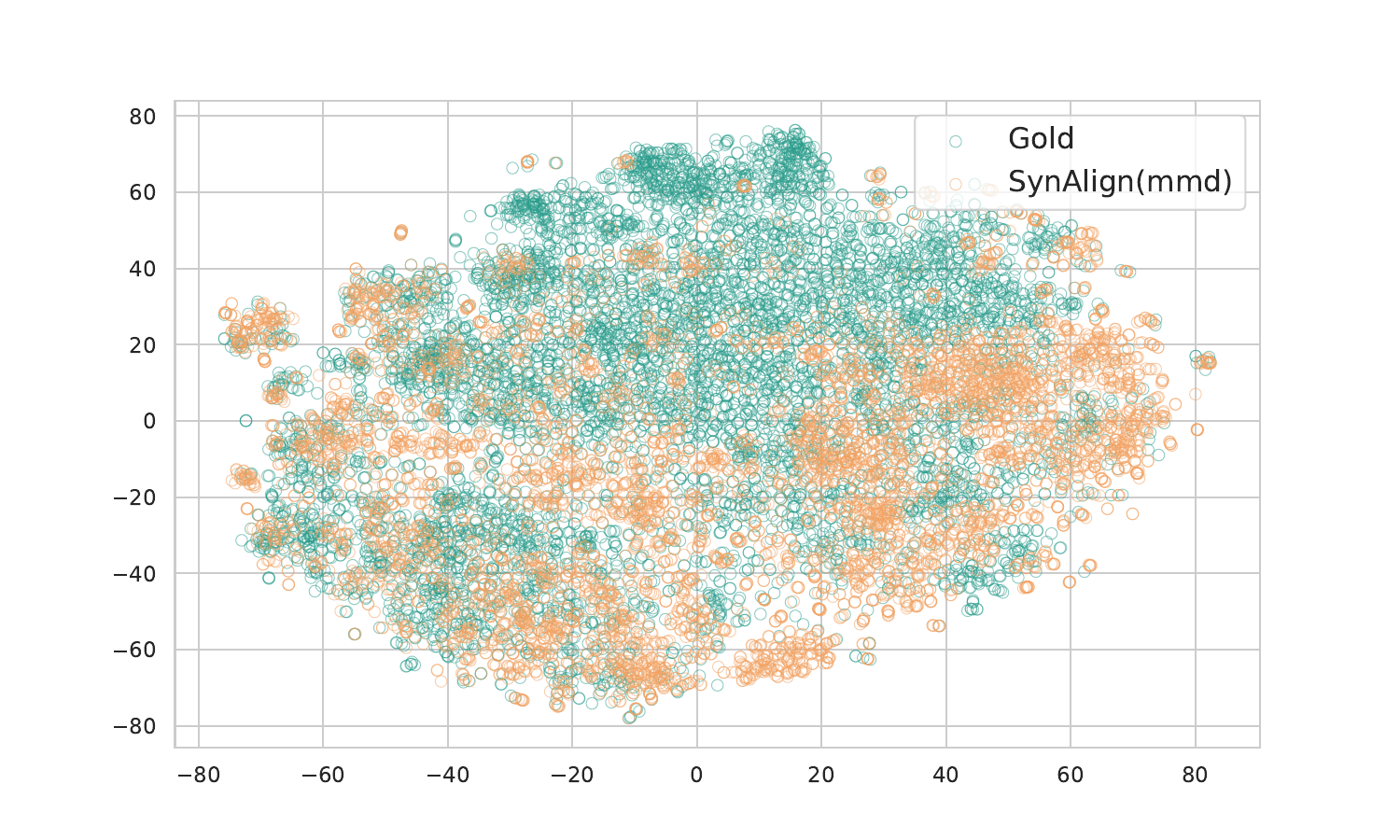}} 
    \caption{t-SNE visualization of sentence embeddings from SST-2. (a) Comparison between Gold and AttrPrompt; (b) Comparison between Gold and SynAlign (MMD). SynAlign (MMD) achieves better alignment.}
    \label{fig:good_distribution}
\end{figure}

Figure~\ref{fig:good_distribution} illustrates the t-SNE visualizations for SST-2, comparing the original dataset (\textbf{Gold}) with data generated by different methods.
In Figure~\ref{fig:good_distribution}(a), AttrPrompt's generated data forms tight clusters far from the original data, suggesting that it captures only a limited subset of the original linguistic patterns.
In contrast, Figure~\ref{fig:good_distribution}(b) shows that \textbf{SynAlign (MMD sampling)} generates data that closely aligns with the original distribution, covering a broader and more representative area of the original data space.

Quantitatively, Table~\ref{tab:was_dis} reports the Wasserstein distances for all datasets. \textbf{SynAlign (random)} achieves lower distances compared to \textbf{SimPrompt} and \textbf{AttrPrompt}, while \textbf{SynAlign (MMD)} consistently achieves the best alignment, demonstrating its effectiveness in selecting synthetic samples that better reflect the original data distribution.

\begin{table}[!t]
\centering
\caption{Wasserstein Distance between original and generated datasets. Lower values indicate better alignment.}
\label{tab:was_dis}
\resizebox{\columnwidth}{!}{
\begin{tabular}{l|c|c|c}
    \hline
    \textbf{Data to compare} & \textbf{SST-2} & \textbf{AGNEWS} & \textbf{Amazon} \\
    \hline
    \textbf{Gold vs SimPrompt} & 0.1924 & 0.0468 & 0.0933 \\
    \textbf{Gold vs AttrPrompt} & 0.0366 & 0.0913 & 0.1640 \\
    \textbf{Gold vs SynAlign(random)} & 0.0082 & 0.0647 & 0.1209 \\
    \rowcolor{gray!30} 
    \textbf{Gold vs SynAlign(mmd)} & \textbf{0.0077} & \textbf{0.0516} & \textbf{0.1068} \\
    \hline
\end{tabular}
}
\end{table}

\noindent \textbf{Vocabulary Diversity:}  
We measure vocabulary diversity by calculating the vocabulary size, defined as the number of unique words in each dataset. As shown in Table~\ref{tab:voc_size}, \textbf{SynAlign (MMD)} generates datasets with higher vocabulary diversity than \textbf{SimPrompt} and comparable diversity to \textbf{AttrPrompt}. For example, on SST-2, SynAlign (MMD) achieves a vocabulary size of 7.4k, significantly larger than SimPrompt (1k) and close to AttrPrompt (7.2k).

Although the generated datasets have smaller vocabulary sizes than the original datasets, this is expected due to the limited input prompts provided to the LLM. However, combining synthetic and original data compensates for this limitation, as evidenced by the performance improvements in Section \ref{sec_method}. These results show that SynAlign not only enhances distributional alignment but also generates more linguistically diverse text, contributing to its superior performance.

\begin{table}[tbp]
\centering
\caption{Vocabulary sizes of original datasets and synthetic datasets generated by different methods.}
\label{tab:voc_size}
\resizebox{0.9\columnwidth}{!}{
\begin{tabular}{l|c|c|c}
    \hline
    \textbf{Data to compare} & \textbf{SST-2} & \textbf{AGNEWS} & \textbf{Amazon} \\
    \hline
    \textbf{Gold} & $13k$ & $158k$ & $90k$ \\
    \textbf{SimPrompt} & $1k$ & $27k$ & $28k$ \\
    \textbf{AttrPrompt} & $7.2k$ & $30k$ & $22k$ \\
    \textbf{SynAlign(all)} & $7.9k$ & $32k$ & $35k$ \\
    \textbf{SynAlign(mmd)} & $7.4k$ & $31k$ & $29k$ \\
    \hline
\end{tabular}
}
\end{table}


\subsection{Ablation Studies}


\subsubsection{Exploration-aware Sampling Coverage Speed}
\label{sec:abla_coverage}


Efficiently covering the original data distribution is critical for few-shot prompting. We compare our \textbf{exploration-aware sampling} with random sampling by evaluating their respective coverage rates of the original data distribution. The coverage area is measured as the convex hull formed by selected samples in the t-SNE-reduced embedding space, with coverage rates computed iteratively for 200 sampling steps.



\begin{figure}[tbp]
    \centering
    \subfigure{\includegraphics[width=0.23\textwidth]{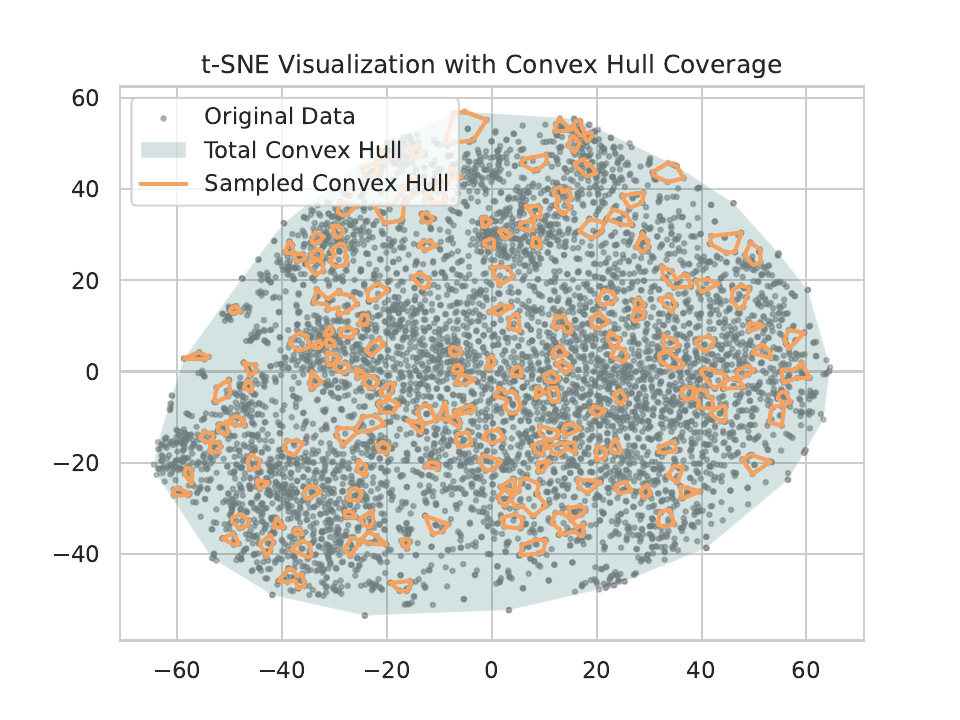}} 
    \subfigure{\includegraphics[width=0.23\textwidth]{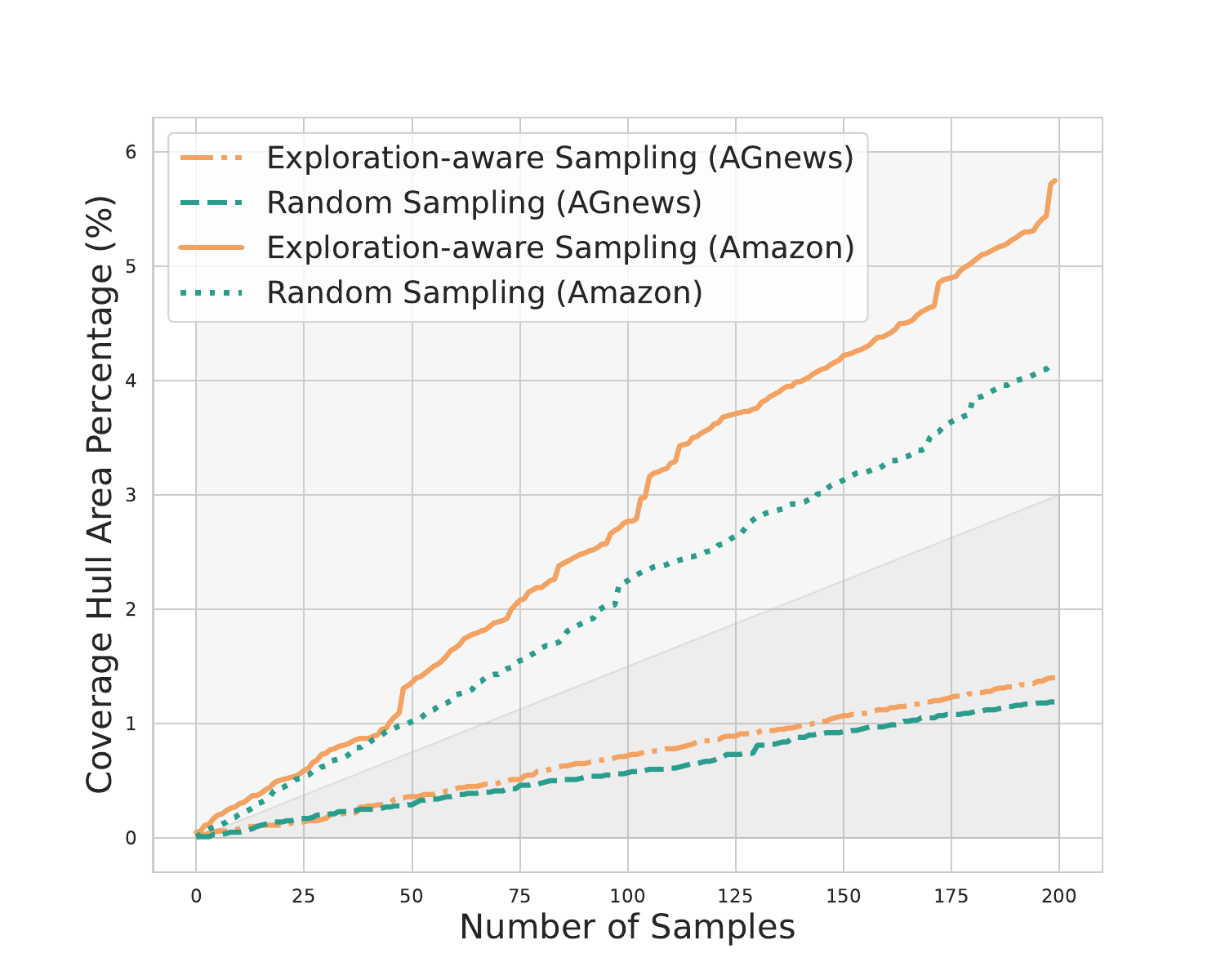}} 
    \caption{(a) Convex hull coverage example for SST-2. (b) Coverage rate comparison between Gaussian Process Active Sampling and random sampling on AGNEWS and Amazon. Gaussian Process achieves faster and broader coverage.}
    \label{fig:gp_cover}
\end{figure}

Figure~\ref{fig:gp_cover}(a) illustrates an example of convex hull coverage for SST-2, while Figure~\ref{fig:gp_cover}(b) compares coverage rates across AGNEWS and Amazon. Exploration-aware sampling consistently achieves faster and broader coverage of the data distribution compared to random sampling. Specifically, it avoids redundant sampling in densely populated regions and captures diverse, representative samples more effectively. This result highlights the efficiency of Gaussian Process Active Sampling in improving few-shot prompting performance.


\subsubsection{Benefits of Latent Attribute Reasoning}

Our method employs a \textbf{two-stage generation process} that separates attribute generation (e.g., sentiment or topic) from synthetic data generation. This design encourages diversity compared to single-stage generation, which directly relies on prompt examples.

Table~\ref{tab:2stage_res} compares the performance of models trained on data generated by single-stage and two-stage approaches. The results show that the two-stage generation consistently outperforms the single-stage approach across AGNEWS and Amazon datasets. For example, on AGNEWS, SynAlign(mmd) achieves an accuracy of 83.81\% with two-stage generation, compared to 83.19\% with single-stage generation. These improvements highlight the importance of reasoning about latent attributes to improve the quality and diversity of synthetic data.

\begin{table}[!t]
\centering
\caption{Performance comparison between single-stage and two-stage generation on AGNEWS and Amazon datasets.}
\label{tab:2stage_res}
\resizebox{\columnwidth}{!}{
\begin{tabular}{ll|c|c}
    \hline
    \textbf{} & \textbf{Method} & \textbf{AGNEWS} & \textbf{Amazon} \\
    \hline
     & \textbf{Gold} & 0.8204 & 0.9353 \\
     & \textbf{AttrPrompt} & 0.8274 & 0.9441 \\
    \hline
    \multirow{3}{*}{\textbf{One Stage}} & \textbf{SynAlign(all)} & 0.8319 & 0.9430 \\
    &\textbf{SynAlign(random)} & 0.8177 & 0.9437 \\
    &\textbf{SynAlign(mmd)} & 0.8319 & 0.9439 \\
    \hline
    \multirow{3}{*}{\textbf{Two Stage}} & \textbf{SynAlign(all)} & 0.8372 & 0.9460 \\
    &\textbf{SynAlign(random)} & 0.8212 & 0.9439 \\
    \rowcolor{gray!30} 
    &\textbf{SynAlign(mmd)} & \textbf{0.8381} & \textbf{0.9475} \\
    \hline
\end{tabular}
}
\end{table}

\subsubsection{Impact of MMD Distribution Alignment}

The MMD-based sampling strategy selects synthetic samples that are closely aligned with the original data distribution. As shown in Table~\ref{tab:main_result}, SynAlign (MMD) achieves the highest performance across all datasets, outperforming both random sampling and other baselines. For example, on AGNEWS, SynAlign (MMD) achieves an accuracy of 0.9475, compared to 0.9429 for random sampling.

The Wasserstein distance results in Table~\ref{tab:was_dis} further validate the effectiveness of MMD sampling. SynAlign (MMD) consistently achieves smaller distances compared to random sampling, indicating better alignment with the original data distribution. This improved alignment explains the observed gains in downstream performance.

\subsection{Hyperparameter Analysis}


\subsubsection{RBF Kernel Length Scale $\tau$ in Exploration-aware Sampling}

The RBF kernel's length scale $\tau$ is a crucial parameter in the Gaussian Process model for tracking sample uncertainty. It controls the smoothness of the covariance function and determines the range of influence of selected samples. Smaller $\tau$ values lead to highly localized effects, while larger values smooth the uncertainty estimates over broader regions.

\begin{figure*}[htbp]
    \centering
    \subfigure{\includegraphics[width=0.24\textwidth]{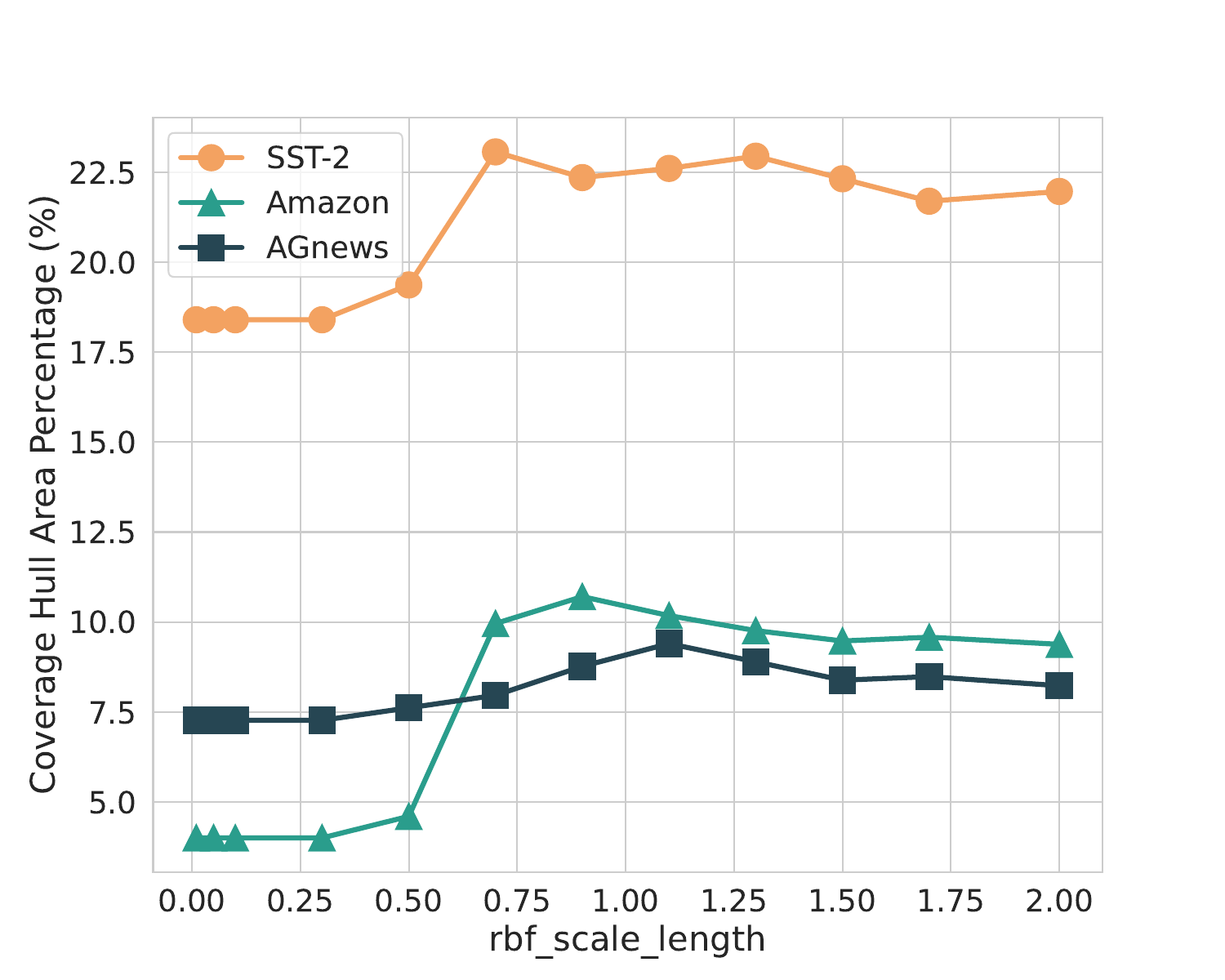}} 
    \subfigure{\includegraphics[width=0.24\textwidth]{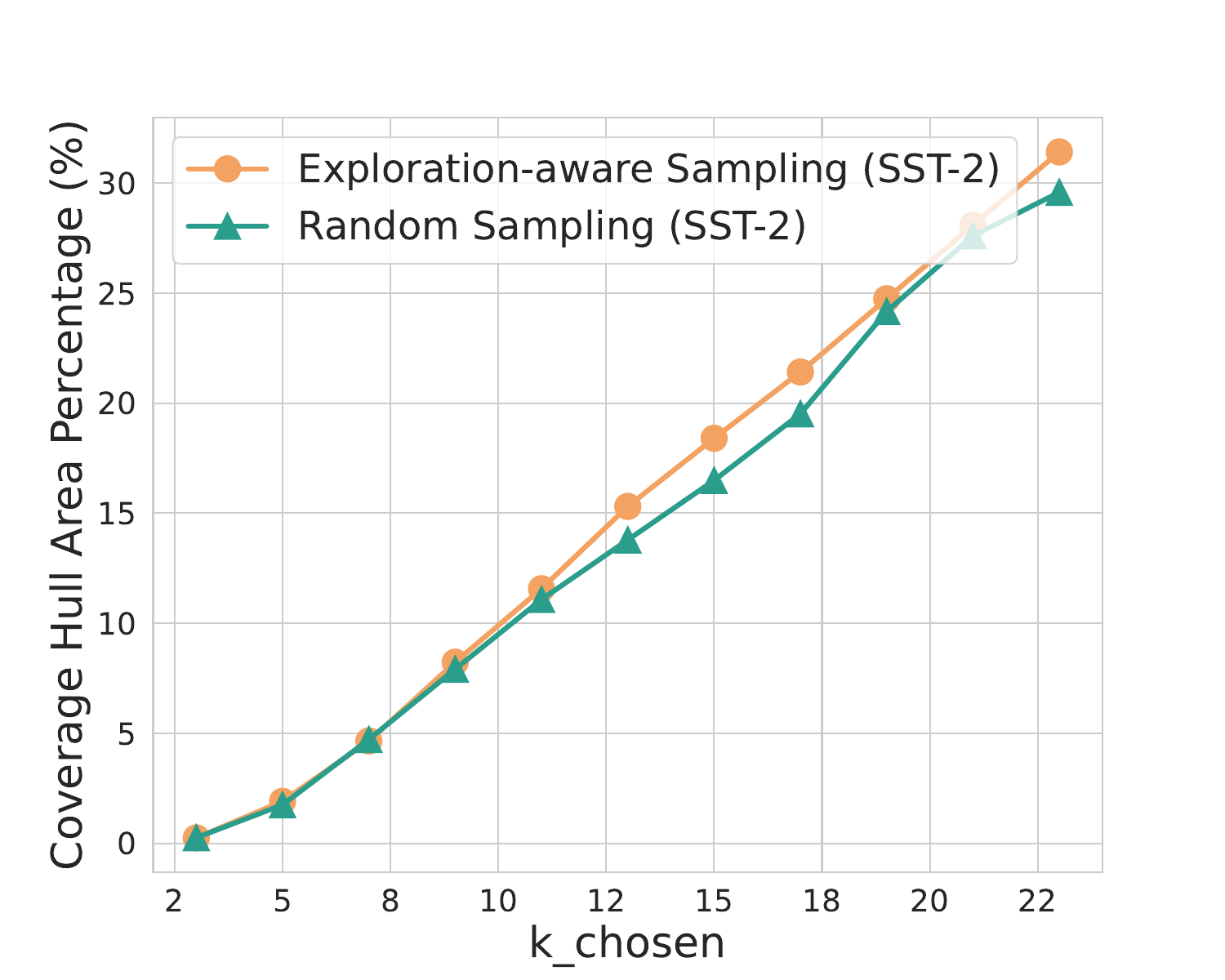}} 
    \subfigure{\includegraphics[width=0.24\textwidth]{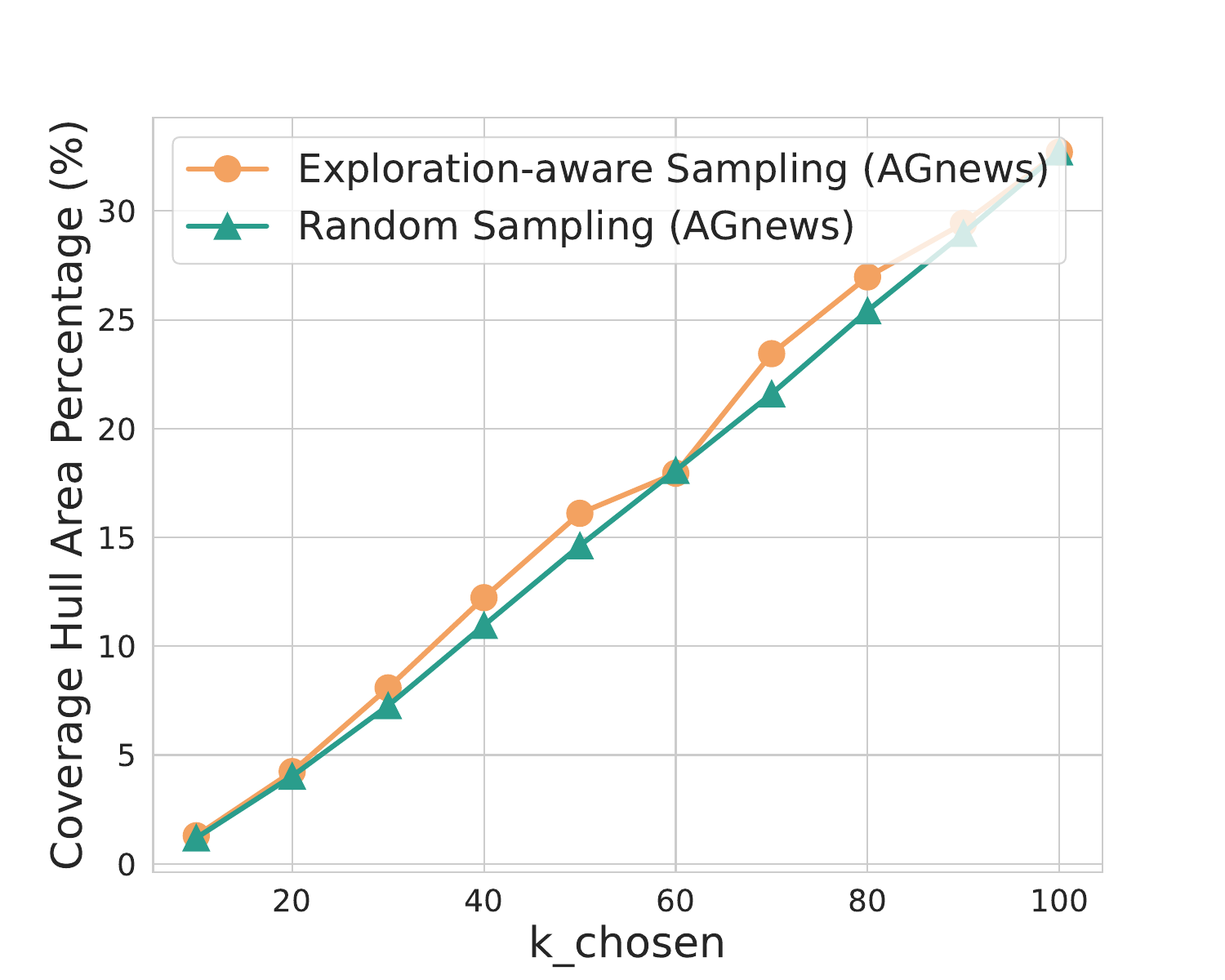}} 
    \subfigure{\includegraphics[width=0.24\textwidth]{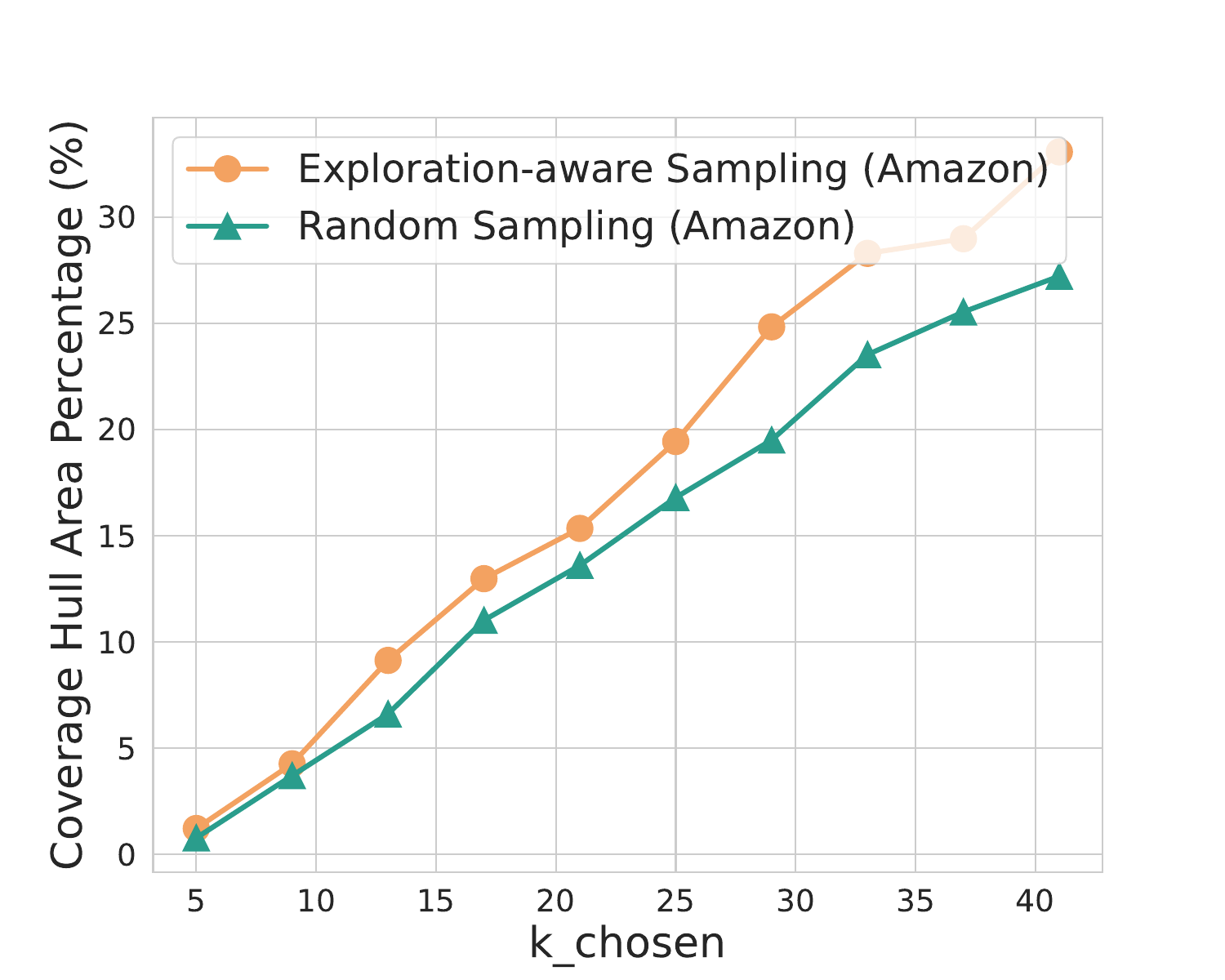}}
    \caption{Coverage rate as a function of RBF kernel length scale $\tau$ for Exploration-aware Sampling across three datasets: SST-2, AGNEWS, and Amazon. Each curve represents the convex hull coverage rate at fixed sampling iterations.}
    \label{fig:para_anal}
\end{figure*}

We evaluate the impact of $\tau$ on the convex hull coverage rate across SST-2, AGNEWS, and Amazon datasets. As shown in Figure~\ref{fig:para_anal} (a), the coverage rate exhibits consistent patterns across datasets. When $\tau$ is too small(e.g., $\tau < 0.3$), the coverage rate is low due to excessive focus on densely populated regions, leading to redundant sampling. As $\tau$ increases, the coverage rate improves, reaching its peak at dataset-specific optimal values (e.g., $\tau = 0.9$ for SST-2). However, when $\tau$ becomes too large ($\tau > 1.5$), the coverage rate declines as the sampling behavior becomes overly smooth, resembling random sampling. 


To better understand the differences between datasets, we analyzed the distribution of pairwise Euclidean distances between sentence embeddings. These analyses, presented in the Appendix, indicate that AGNEWS has a more spread-out embedding space compared to SST-2 and Amazon, which explains why it benefits from a larger $\tau$ for optimal coverage.

\subsubsection{Number of Nearest Neighbors $k$ in Exploration-aware Sampling}

The parameter $k$, which determines the number of nearest neighbors selected during each sampling iteration, controls the trade-off between sample diversity and efficiency in Exploration-aware Sampling. Larger $k$ values enable broader coverage of the embedding space, while smaller values focus on fewer, more representative samples.


We evaluate the impact of $k$ on the convex hull coverage rate under the optimal RBF kernel length scale identified earlier. Figure~\ref{fig:para_anal} (b)-(d) shows that the coverage rate increases with $k$ across all datasets, but the rate of improvement diminishes as $k$ becomes large.
However, larger $k$ values incur higher computational costs, particularly for datasets with large embedding spaces like AGNEWS.
Interestingly, when $k$ is small, the performance of Exploration-aware Sampling is comparable to random sampling. 
This demonstrates that our method is more effective at utilizing larger sampling budgets to achieve higher coverage.


\subsubsection{Number of Projection Matrices $|\Theta|$ in Synthetic Distribution Alignment}

In the Synthetic Distribution Alignment stage, the number of projection matrices $|\Theta|$ plays a crucial role in aligning the original data distribution $D_{ori}$ with the generated data distribution $D_{gen}$ through MMD. This parameter determines the expressiveness of the alignment process: too few projection matrices may inadequately capture distributional differences, while too many may lead to computational overhead or overfitting.

To evaluate the effect of $|\Theta|$, we vary its value across \{10, 50, 100, 500, 1000\} and measure the Wasserstein distance between $D_{ori}$ and $D_{gen}$. Table~\ref{tab:theta_analysis} reports the results for SST-2, AGNEWS, and Amazon.
For SST-2 and Amazon, the Wasserstein distance is minimized at $|\Theta| = 50$, indicating that moderate numbers of projection matrices are sufficient for effective alignment in datasets with simpler embedding spaces. In contrast, for AGNEWS, which has a more complex embedding space due to its higher diversity, the Wasserstein distance continues to decrease as $|\Theta|$ increases, reaching its lowest value at $|\Theta| = 100$ (0.00715). Further increases in $|\Theta|$ yield diminishing returns while increasing computational costs.


\begin{table}[htbp]
\centering
\caption{Wasserstein distance between $D_{ori}$ and $D_{gen}$ as a function of the number of projection matrices $|\Theta|$.}
\label{tab:theta_analysis}
\begin{tabular}{lccccc}
\toprule
\textbf{Dataset} & \textbf{$10$} & \textbf{$50$} & \textbf{$100$} & \textbf{$500$} & \textbf{$1000$} \\
\midrule
SST-2 & 0.11069 & \textbf{0.10516} & 0.10735 & 0.10931 & 0.10801 \\
AGNEWS & 0.00773 & 0.00748 & \textbf{0.00715} & 0.00705 & 0.00702 \\
Amazon & 0.06055 & \textbf{0.06357} & 0.06419 & 0.06375 & 0.06310 \\
\bottomrule
\end{tabular}
\end{table}

\subsection{Online A/B Test}

We deployed the SynAlign framework in the pre-ranking module of AppGallery's search advertising system to address two key challenges: the long-tail nature of app ads, which results in limited search data, and the significant query-app discrepancies caused by abstract app names (e.g., 'Presidential Election' vs. 'TikTok'). Training relevance models on exposure-click data using contrastive learning struggles with generalization due to these issues. While LLMs can augment user queries, zero-shot LLM synthesis often generates queries that deviate from real-world user preferences, leading to distribution mismatches and potential model convergence issues.

To address the distribution mismatch between synthetic and real queries, we utilized the SynAlign framework. During the SynAlign synthesis process, we applied uncertainty sampling to efficiently analyze all query-item pairs and identify common user query patterns (e.g., app names, substrings, typos). These patterns were then combined with app names and input into Qwen2.5 to generate hundreds of thousands of synthetic queries. Both synthetic and real queries were mapped into the embedding space, where SynAlign’s MMD-based method was used to assign a weight to each synthetic query. A new dataset was created by sampling queries based on these weights. Offline testing showed that models enhanced with distribution-aligned synthetic queries achieved a 0.26\% improvement in AUC compared to unenhanced models.

For online deployment, synthetic queries were mixed with daily updated real queries for the experimental group, while the control group used only real queries. Over a week, the experimental group achieved a 2.86\% increase in RPM and a 2.31\% increase in CPM. This SynAlign-based data synthesis approach is now fully deployed to serve all users.

%% file: sections/5_conclusion.tex
\section{Conclusion}
In conclusion, we observed that the data generated by LLMs often struggles to align perfectly with domain-specific linguistic styles. Directly mixing LLM-generated data with original data can disrupt the original data distribution, leading to degraded model performance. To address this issue, we proposed the SynAlign framework. This framework begins with an Exploration-aware Sampling module, which allows the LLM to efficiently perceive the full distribution of real-world data. Next, the Latent-Attribute Reasoning module summarizes and generalizes the linguistic attributes to guide the LLM in generating standardized synthetic data. Finally, the Synthetic Distribution Alignment module uses an MMD-based approach to align the distributions of synthetic and original data, effectively enhancing the performance of domain-specific tasks. Extensive experiments were conducted and proved the effectiveness of our method.


%% file: sections/6_appendix.tex


\section{Appendix}

\begin{algorithm}[bp]
	\caption{The Algorithm of the Proposed SynAlign} 
	\label{apdx:alg} 
	\begin{algorithmic}
		\STATE \textbf{Input}: Real dataset $D_{ori}$, Sample uncertainty tracker $U(\cdot)$, Key Attribute Set $A$, Embedding model $F$
		\STATE \textbf{Output}: Original Dataset $D_{gen}$
		\STATE Mapping $D_{ori}$ to $E_{ori}$ with $F$
		\STATE Initialize $U(E_{ori})$ with standard GP model with RBF kernel 
		\STATE Initialize generated text pool $D_{gen}=\emptyset$
		\WHILE{$max(U(E_{ori})) > \sigma$}  
		\STATE Select demonstrations $D_{dem}$ with formula 3
		\STATE Set $U(E_{dem})$ as 0 and update $U(E_{ori})$ with formula 4, 5
		\STATE Extract key attribute set $S$ from $D_{dem}$ with formula 6
		\STATE Generate $D_{gen}^{'}$ based on $S$ with formula 7
		\STATE $D_{gen}$ = $D_{gen}^{'}\cup D_{gen}$     
		\ENDWHILE 
		\STATE Mapping $D_{gen}$ to $E_{gen}$ with $F$
		\STATE Initialize Random Matrix set $\Theta$ with Gram-Schmidt algorithm
		\STATE Initialize sampling weight $\omega$ for each embedding in $E_{gen}$
		\STATE Train $\omega$ by minimizing MMD loss between $E_{ori}$ and $E_{gen}$
		\STATE Resample $D_{gen}$ based on $\omega$ as the final synthetic data $D_{gen}$
		\STATE \textbf{Return} $D_{gen}$
	\end{algorithmic} 
	\label{alg1}
\end{algorithm}
\vspace*{-0.5cm}

\subsection{Dataset Information}
\label{apdx:data_info}

To evaluate the effectiveness of SynAlign, we conduct experiments on three widely used datasets: SST-2, AGNEWS, and Amazon. Table~\ref{tab:data_info} summarizes the key attributes of these datasets.


For each dataset, we report the number of training samples, test samples, and classes. Additionally, we include the amount of synthetic data generated by the LLM and the final number of samples selected through SynAlign's sampling strategy. These datasets span diverse domains (e.g., reviews, news, and web content), ensuring a comprehensive evaluation of the proposed framework.

\begin{table}[bp]
	\centering
	\caption{Summary of datasets used in the experiments.}
	\label{tab:data_info}
	\begin{tabular}{lccc}
		\toprule
		\textbf{Attribute} & \textbf{SST-2} & \textbf{AGNEWS} & \textbf{Amazon} \\
		\midrule
		\textbf{Domain} & Review & News & Web \\
		\textbf{\#Train} & $67k$ & $120k$ & $13.8k$ \\
		\textbf{\#Test} & $1.8k$ & $7.6k$ & $1.2k$ \\
		\textbf{\#Class} & 2 & 4 & 23 \\
		\textbf{\#Generated} & $9k$ & $16k$ & $18.4k$ \\
		\textbf{\#Sampled} & $6k$ & $6k$ & $13.8k$ \\
		\bottomrule
	\end{tabular}
\end{table}

\subsection{Implementation Details}
\label{apdx:imp_detail}

\subsubsection{Hardware Configuration}

All experiments were conducted on a server equipped with an NVIDIA A6000 GPU (48GB memory) and an Intel(R) Xeon(R) Gold 6242R CPU @ 3.10GHz.

\subsubsection{Classifier Training Parameters}

For the final classification task, we fine-tune a pre-trained language model using the parameters listed in Table~\ref{tab:classifier_params}. These include the learning rate (\texttt{lr}), batch size, training epochs, weight decay, and warmup ratio. The warmup ratio specifies the proportion of training steps used for learning rate warmup to stabilize training.


\begin{table}[bp]
	\centering
	\caption{Parameters for Classifier Training.}
	\label{tab:classifier_params}
	\begin{tabular}{lccc}
		\toprule
		\textbf{Parameter} & \textbf{SST-2} & \textbf{AGNEWS} & \textbf{Amazon} \\
		\midrule
		\textbf{lr} & 5e-5 & 5e-5 & 5e-5 \\
		\textbf{Batch size} & 32 & 32 & 32 \\
		\textbf{Training epochs} & 6 & 6 & 6 \\
		\textbf{Weight decay} & 1e-4 & 1e-4 & 1e-4 \\
		\textbf{Warmup ratio} & 6\% & 6\% & 6\% \\
		\bottomrule
	\end{tabular}
\end{table}


\subsubsection{Convex Hull Coverage}
\label{apdx:convex}

To calculate the coverage rate in \ref{sec:abla_coverage}, we first reduce the dimensionality of the sentence embeddings for both the original and generated datasets using t-SNE. A k-d tree is then constructed using the 2D t-SNE embeddings of the original dataset to enable efficient neighbor searching. The convex hull of the t-SNE embeddings of the original dataset is taken as the target distribution's total coverage area. We initialize an empty buffer set $\mathcal{B}$ to store convex hulls formed during sampling. At each sampling iteration, a single example is selected, and its embedding is combined with the $k$-nearest neighbors (identified using the k-d tree) to create a new convex hull. If this convex hull overlaps with any existing convex hulls in $\mathcal{B}$, they are merged to form a larger convex hull. The total area of all convex hulls in $\mathcal{B}$ is then calculated and compared to the total coverage area of the original dataset to compute the coverage rate. This process is repeated iteratively for 200 sampling steps.




\subsection{Prompt Design}

\begin{figure}[tbp]
	\centering
	\includegraphics[width=0.45\textwidth]{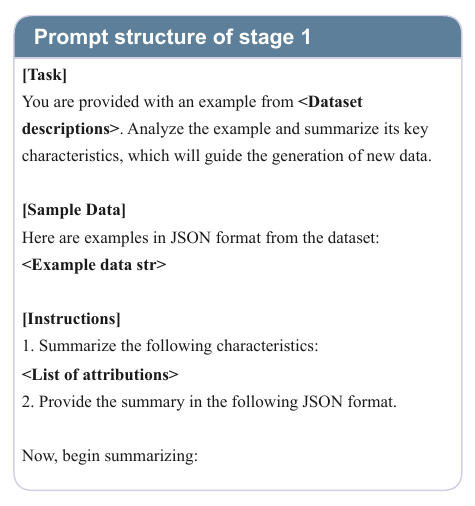}
	\caption{Prompt used in Stage 1 (Attribute Summarization). The LLM is instructed to analyze a dataset example and extract key linguistic and semantic attributes, such as topics, language habits, and writing styles, to guide the synthetic data generation process.}
	\label{fig:prompt_s1}
\end{figure}

\begin{figure}[bp]
	\centering
	\includegraphics[width=0.45\textwidth]{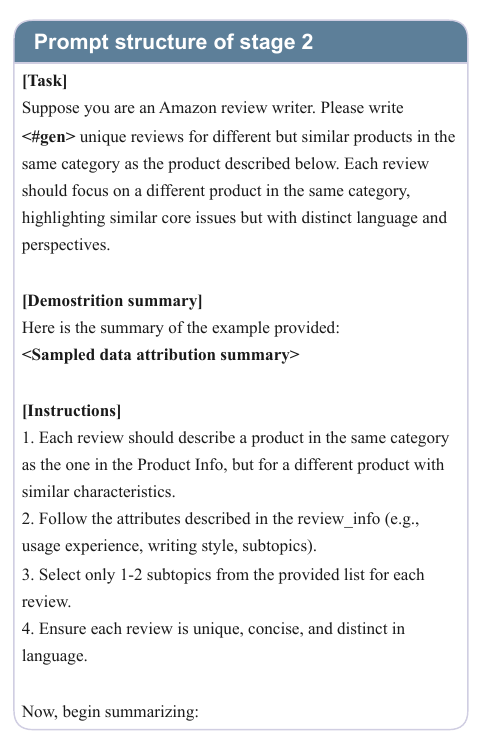}
	\caption{Prompt used in Stage 2 (Synthetic Data Generation). The LLM generates multiple unique samples based on the attributes summarized in Stage 1. For example, in the Amazon dataset, it generates reviews for similar products while ensuring diversity in subtopics and language styles.}
	\label{fig:prompt_s2}
\end{figure}

To guide the synthetic data generation process, we design two stages of prompts tailored to capture and generalize the linguistic attributes of each dataset. The prompts are presented as examples in Figures~\ref{fig:prompt_s1} and \ref{fig:prompt_s2}.

\subsubsection{Stage 1: Attribute Summarization}

In the first stage, the LLM analyzes a set of example data from the target dataset and generates a structured summary of its key linguistic and semantic attributes. The prompt instructs the model to extract attributes such as topics, language habits, and writing styles, depending on the dataset. For instance, for the SST-2 dataset, the attributes include \textit{movie genres}, \textit{topics}, \textit{language habits}, and \textit{review length}. This summarization step ensures that the generated synthetic data adheres to the original dataset’s characteristics.

\subsubsection{Stage 2: Synthetic Data Generation}

In the second stage, the LLM generates synthetic data based on the summarized attributes from the first stage. For example, in the Amazon dataset, the model is instructed to write multiple unique product reviews in the same category while adhering to the provided attribute summary. The prompt ensures diversity by specifying that each review should focus on distinct subtopics and employ varied language styles. Similar prompts are designed for the SST-2 and AGNEWS datasets, with dataset-specific attributes and generation instructions.

\subsubsection{Dataset-specific Attributes}

The attributes extracted and used for generation vary across datasets, as summarized below:

\begin{itemize}
	\item \textbf{SST-2:} Movie genres, topics, language habits, and review length.
	\item \textbf{AGNEWS:} News topics, writing style, news length, subtopics, and location.
	\item \textbf{Amazon:} Product information, usage experience, writing style, review length, language habits, and subtopics.
\end{itemize}

These prompts are designed to adapt to the characteristics of each dataset, ensuring that the synthetic data captures the essential features of the original data while maintaining diversity and linguistic consistency.

%% file: main.bbl

\begin{thebibliography}{56}


\ifx \showCODEN    \undefined \def \showCODEN     #1{\unskip}     \fi
\ifx \showDOI      \undefined \def \showDOI       #1{#1}\fi
\ifx \showISBNx    \undefined \def \showISBNx     #1{\unskip}     \fi
\ifx \showISBNxiii \undefined \def \showISBNxiii  #1{\unskip}     \fi
\ifx \showISSN     \undefined \def \showISSN      #1{\unskip}     \fi
\ifx \showLCCN     \undefined \def \showLCCN      #1{\unskip}     \fi
\ifx \shownote     \undefined \def \shownote      #1{#1}          \fi
\ifx \showarticletitle \undefined \def \showarticletitle #1{#1}   \fi
\ifx \showURL      \undefined \def \showURL       {\relax}        \fi
\providecommand\bibfield[2]{#2}
\providecommand\bibinfo[2]{#2}
\providecommand\natexlab[1]{#1}
\providecommand\showeprint[2][]{arXiv:#2}

\bibitem[Arora et~al\mbox{.}(2022)]%
        {arora2022exposure}
\bibfield{author}{\bibinfo{person}{Kushal Arora}, \bibinfo{person}{Layla~El
  Asri}, \bibinfo{person}{Hareesh Bahuleyan}, {and} \bibinfo{person}{Jackie
  Chi~Kit Cheung}.} \bibinfo{year}{2022}\natexlab{}.
\newblock \showarticletitle{Why exposure bias matters: An imitation learning
  perspective of error accumulation in language generation}.
\newblock \bibinfo{journal}{\emph{arXiv preprint arXiv:2204.01171}}
  (\bibinfo{year}{2022}).
\newblock


\bibitem[Blitzer et~al\mbox{.}(2007)]%
        {amazon}
\bibfield{author}{\bibinfo{person}{John Blitzer}, \bibinfo{person}{Mark
  Dredze}, {and} \bibinfo{person}{Fernando Pereira}.}
  \bibinfo{year}{2007}\natexlab{}.
\newblock \showarticletitle{Biographies, bollywood, boom-boxes and blenders:
  Domain adaptation for sentiment classification}. In
  \bibinfo{booktitle}{\emph{Proceedings of the 45th annual meeting of the
  association of computational linguistics}}. \bibinfo{pages}{440--447}.
\newblock


\bibitem[Brown et~al\mbox{.}(2020)]%
        {brown2020language}
\bibfield{author}{\bibinfo{person}{Tom Brown}, \bibinfo{person}{Benjamin Mann},
  \bibinfo{person}{Nick Ryder}, \bibinfo{person}{Melanie Subbiah},
  \bibinfo{person}{Jared~D Kaplan}, \bibinfo{person}{Prafulla Dhariwal},
  \bibinfo{person}{Arvind Neelakantan}, \bibinfo{person}{Pranav Shyam},
  \bibinfo{person}{Girish Sastry}, \bibinfo{person}{Amanda Askell},
  {et~al\mbox{.}}} \bibinfo{year}{2020}\natexlab{}.
\newblock \showarticletitle{Language models are few-shot learners}.
\newblock \bibinfo{journal}{\emph{Advances in neural information processing
  systems}}  \bibinfo{volume}{33} (\bibinfo{year}{2020}),
  \bibinfo{pages}{1877--1901}.
\newblock


\bibitem[Chen et~al\mbox{.}(2023)]%
        {chen2023mixturesoftpromptscontrollable}
\bibfield{author}{\bibinfo{person}{Derek Chen}, \bibinfo{person}{Celine Lee},
  \bibinfo{person}{Yunan Lu}, \bibinfo{person}{Domenic Rosati}, {and}
  \bibinfo{person}{Zhou Yu}.} \bibinfo{year}{2023}\natexlab{}.
\newblock \bibinfo{title}{Mixture of Soft Prompts for Controllable Data
  Generation}.
\newblock
\newblock
\showeprint[arxiv]{2303.01580}~[cs.CL]
\urldef\tempurl%
\url{https://arxiv.org/abs/2303.01580}
\showURL{%
\tempurl}


\bibitem[Dai et~al\mbox{.}(2024a)]%
        {dai2024cocktail}
\bibfield{author}{\bibinfo{person}{Sunhao Dai}, \bibinfo{person}{Weihao Liu},
  \bibinfo{person}{Yuqi Zhou}, \bibinfo{person}{Liang Pang},
  \bibinfo{person}{Rongju Ruan}, \bibinfo{person}{Gang Wang},
  \bibinfo{person}{Zhenhua Dong}, \bibinfo{person}{Jun Xu}, {and}
  \bibinfo{person}{Ji-Rong Wen}.} \bibinfo{year}{2024}\natexlab{a}.
\newblock \showarticletitle{Cocktail: A Comprehensive Information Retrieval
  Benchmark with LLM-Generated Documents Integration}.
\newblock \bibinfo{journal}{\emph{Findings of the Association for Computational
  Linguistics: ACL 2024}} (\bibinfo{year}{2024}).
\newblock


\bibitem[Dai et~al\mbox{.}(2023b)]%
        {dai2023uncovering}
\bibfield{author}{\bibinfo{person}{Sunhao Dai}, \bibinfo{person}{Ninglu Shao},
  \bibinfo{person}{Haiyuan Zhao}, \bibinfo{person}{Weijie Yu},
  \bibinfo{person}{Zihua Si}, \bibinfo{person}{Chen Xu},
  \bibinfo{person}{Zhongxiang Sun}, \bibinfo{person}{Xiao Zhang}, {and}
  \bibinfo{person}{Jun Xu}.} \bibinfo{year}{2023}\natexlab{b}.
\newblock \showarticletitle{Uncovering chatgpt’s capabilities in recommender
  systems}. In \bibinfo{booktitle}{\emph{Proceedings of the 17th ACM Conference
  on Recommender Systems}}. \bibinfo{pages}{1126--1132}.
\newblock


\bibitem[Dai et~al\mbox{.}(2024b)]%
        {dai2024modeling}
\bibfield{author}{\bibinfo{person}{Sunhao Dai}, \bibinfo{person}{Ninglu Shao},
  \bibinfo{person}{Jieming Zhu}, \bibinfo{person}{Xiao Zhang},
  \bibinfo{person}{Zhenhua Dong}, \bibinfo{person}{Jun Xu},
  \bibinfo{person}{Quanyu Dai}, {and} \bibinfo{person}{Ji-Rong Wen}.}
  \bibinfo{year}{2024}\natexlab{b}.
\newblock \showarticletitle{Modeling user attention in music recommendation}.
  In \bibinfo{booktitle}{\emph{2024 IEEE 40th International Conference on Data
  Engineering (ICDE)}}. IEEE, \bibinfo{pages}{761--774}.
\newblock


\bibitem[Dai et~al\mbox{.}(2023c)]%
        {dai2023dually}
\bibfield{author}{\bibinfo{person}{Sunhao Dai}, \bibinfo{person}{Yuqi Zhou},
  \bibinfo{person}{Jun Xu}, {and} \bibinfo{person}{Ji-Rong Wen}.}
  \bibinfo{year}{2023}\natexlab{c}.
\newblock \showarticletitle{Dually Enhanced Delayed Feedback Modeling for
  Streaming Conversion Rate Prediction}. In
  \bibinfo{booktitle}{\emph{Proceedings of the 32nd ACM International
  Conference on Information and Knowledge Management}}.
  \bibinfo{pages}{390--399}.
\newblock


\bibitem[Dai et~al\mbox{.}(2023a)]%
        {dai2023long}
\bibfield{author}{\bibinfo{person}{Yi Dai}, \bibinfo{person}{Hao Lang},
  \bibinfo{person}{Yinhe Zheng}, \bibinfo{person}{Fei Huang}, {and}
  \bibinfo{person}{Yongbin Li}.} \bibinfo{year}{2023}\natexlab{a}.
\newblock \showarticletitle{Long-tailed question answering in an open world}.
\newblock \bibinfo{journal}{\emph{arXiv preprint arXiv:2305.06557}}
  (\bibinfo{year}{2023}).
\newblock


\bibitem[Daniel et~al\mbox{.}(2018)]%
        {daniel2018quality}
\bibfield{author}{\bibinfo{person}{Florian Daniel}, \bibinfo{person}{Pavel
  Kucherbaev}, \bibinfo{person}{Cinzia Cappiello}, \bibinfo{person}{Boualem
  Benatallah}, {and} \bibinfo{person}{Mohammad Allahbakhsh}.}
  \bibinfo{year}{2018}\natexlab{}.
\newblock \showarticletitle{Quality control in crowdsourcing: A survey of
  quality attributes, assessment techniques, and assurance actions}.
\newblock \bibinfo{journal}{\emph{ACM Computing Surveys (CSUR)}}
  \bibinfo{volume}{51}, \bibinfo{number}{1} (\bibinfo{year}{2018}),
  \bibinfo{pages}{1--40}.
\newblock


\bibitem[Devlin et~al\mbox{.}(2019)]%
        {bert}
\bibfield{author}{\bibinfo{person}{Jacob Devlin}, \bibinfo{person}{Ming-Wei
  Chang}, \bibinfo{person}{Kenton Lee}, {and} \bibinfo{person}{Kristina
  Toutanova}.} \bibinfo{year}{2019}\natexlab{}.
\newblock \bibinfo{title}{BERT: Pre-training of Deep Bidirectional Transformers
  for Language Understanding}.
\newblock
\newblock
\showeprint[arxiv]{1810.04805}~[cs.CL]
\urldef\tempurl%
\url{https://arxiv.org/abs/1810.04805}
\showURL{%
\tempurl}


\bibitem[Du et~al\mbox{.}(2024a)]%
        {du2024lightcs}
\bibfield{author}{\bibinfo{person}{Zhaocheng Du}, \bibinfo{person}{Junhao
  Chen}, \bibinfo{person}{Qinglin Jia}, \bibinfo{person}{Chuhan Wu},
  \bibinfo{person}{Jieming Zhu}, \bibinfo{person}{Zhenhua Dong}, {and}
  \bibinfo{person}{Ruiming Tang}.} \bibinfo{year}{2024}\natexlab{a}.
\newblock \showarticletitle{LightCS: Selecting Quadratic Feature Crosses in
  Linear Complexity}. In \bibinfo{booktitle}{\emph{Companion Proceedings of the
  ACM on Web Conference 2024}}. \bibinfo{pages}{38--46}.
\newblock


\bibitem[Du et~al\mbox{.}(2024b)]%
        {du2024tutorial}
\bibfield{author}{\bibinfo{person}{Zhaocheng Du}, \bibinfo{person}{Chuhan Wu},
  \bibinfo{person}{Qinglin Jia}, \bibinfo{person}{Jieming Zhu}, {and}
  \bibinfo{person}{Xu Chen}.} \bibinfo{year}{2024}\natexlab{b}.
\newblock \showarticletitle{A Tutorial on Feature Interpretation in Recommender
  Systems}. In \bibinfo{booktitle}{\emph{Proceedings of the 18th ACM Conference
  on Recommender Systems}}. \bibinfo{pages}{1281--1282}.
\newblock


\bibitem[Eldan and Li(2023)]%
        {tinystories}
\bibfield{author}{\bibinfo{person}{Ronen Eldan} {and} \bibinfo{person}{Yuanzhi
  Li}.} \bibinfo{year}{2023}\natexlab{}.
\newblock \bibinfo{title}{TinyStories: How Small Can Language Models Be and
  Still Speak Coherent English?}
\newblock
\newblock
\showeprint[arxiv]{2305.07759}~[cs.CL]
\urldef\tempurl%
\url{https://arxiv.org/abs/2305.07759}
\showURL{%
\tempurl}


\bibitem[Feng et~al\mbox{.}(2020)]%
        {feng2020genaug}
\bibfield{author}{\bibinfo{person}{Steven~Y Feng}, \bibinfo{person}{Varun
  Gangal}, \bibinfo{person}{Dongyeop Kang}, \bibinfo{person}{Teruko Mitamura},
  {and} \bibinfo{person}{Eduard Hovy}.} \bibinfo{year}{2020}\natexlab{}.
\newblock \showarticletitle{Genaug: Data augmentation for finetuning text
  generators}.
\newblock \bibinfo{journal}{\emph{arXiv preprint arXiv:2010.01794}}
  (\bibinfo{year}{2020}).
\newblock


\bibitem[Gao et~al\mbox{.}(2025)]%
        {gao2025samplellm}
\bibfield{author}{\bibinfo{person}{Jingtong Gao}, \bibinfo{person}{Zhaocheng
  Du}, \bibinfo{person}{Xiaopeng Li}, \bibinfo{person}{Xiangyu Zhao},
  \bibinfo{person}{Yichao Wang}, \bibinfo{person}{Xiangyang Li},
  \bibinfo{person}{Huifeng Guo}, {and} \bibinfo{person}{Ruiming Tang}.}
  \bibinfo{year}{2025}\natexlab{}.
\newblock \showarticletitle{SampleLLM: Optimizing Tabular Data Synthesis in
  Recommendations}.
\newblock \bibinfo{journal}{\emph{arXiv preprint arXiv:2501.16125}}
  (\bibinfo{year}{2025}).
\newblock


\bibitem[Gao et~al\mbox{.}(2023)]%
        {sungen}
\bibfield{author}{\bibinfo{person}{Jiahui Gao}, \bibinfo{person}{Renjie Pi},
  \bibinfo{person}{Yong Lin}, \bibinfo{person}{Hang Xu},
  \bibinfo{person}{Jiacheng Ye}, \bibinfo{person}{Zhiyong Wu},
  \bibinfo{person}{Weizhong Zhang}, \bibinfo{person}{Xiaodan Liang},
  \bibinfo{person}{Zhenguo Li}, {and} \bibinfo{person}{Lingpeng Kong}.}
  \bibinfo{year}{2023}\natexlab{}.
\newblock \bibinfo{title}{Self-Guided Noise-Free Data Generation for Efficient
  Zero-Shot Learning}.
\newblock
\newblock
\showeprint[arxiv]{2205.12679}~[cs.CL]
\urldef\tempurl%
\url{https://arxiv.org/abs/2205.12679}
\showURL{%
\tempurl}


\bibitem[Gilardi et~al\mbox{.}(2023)]%
        {ChatGPTOC}
\bibfield{author}{\bibinfo{person}{Fabrizio Gilardi}, \bibinfo{person}{Meysam
  Alizadeh}, {and} \bibinfo{person}{Ma{\"e}l Kubli}.}
  \bibinfo{year}{2023}\natexlab{}.
\newblock \showarticletitle{ChatGPT outperforms crowd workers for
  text-annotation tasks}.
\newblock \bibinfo{journal}{\emph{Proceedings of the National Academy of
  Sciences of the United States of America}}  \bibinfo{volume}{120}
  (\bibinfo{year}{2023}).
\newblock
\urldef\tempurl%
\url{https://api.semanticscholar.org/CorpusID:257766307}
\showURL{%
\tempurl}


\bibitem[Gretton et~al\mbox{.}(2012)]%
        {gretton2012kernel}
\bibfield{author}{\bibinfo{person}{Arthur Gretton}, \bibinfo{person}{Karsten~M
  Borgwardt}, \bibinfo{person}{Malte~J Rasch}, \bibinfo{person}{Bernhard
  Sch{\"o}lkopf}, {and} \bibinfo{person}{Alexander Smola}.}
  \bibinfo{year}{2012}\natexlab{}.
\newblock \showarticletitle{A kernel two-sample test}.
\newblock \bibinfo{journal}{\emph{The Journal of Machine Learning Research}}
  \bibinfo{volume}{13}, \bibinfo{number}{1} (\bibinfo{year}{2012}),
  \bibinfo{pages}{723--773}.
\newblock


\bibitem[He et~al\mbox{.}(2023)]%
        {he2023annollm}
\bibfield{author}{\bibinfo{person}{Xingwei He}, \bibinfo{person}{Zhenghao Lin},
  \bibinfo{person}{Yeyun Gong}, \bibinfo{person}{Alex Jin},
  \bibinfo{person}{Hang Zhang}, \bibinfo{person}{Chen Lin},
  \bibinfo{person}{Jian Jiao}, \bibinfo{person}{Siu~Ming Yiu},
  \bibinfo{person}{Nan Duan}, \bibinfo{person}{Weizhu Chen}, {et~al\mbox{.}}}
  \bibinfo{year}{2023}\natexlab{}.
\newblock \showarticletitle{Annollm: Making large language models to be better
  crowdsourced annotators}.
\newblock \bibinfo{journal}{\emph{arXiv preprint arXiv:2303.16854}}
  (\bibinfo{year}{2023}).
\newblock


\bibitem[Hsieh et~al\mbox{.}(2023)]%
        {hsieh2023distilling}
\bibfield{author}{\bibinfo{person}{Cheng-Yu Hsieh}, \bibinfo{person}{Chun-Liang
  Li}, \bibinfo{person}{Chih-Kuan Yeh}, \bibinfo{person}{Hootan Nakhost},
  \bibinfo{person}{Yasuhisa Fujii}, \bibinfo{person}{Alexander Ratner},
  \bibinfo{person}{Ranjay Krishna}, \bibinfo{person}{Chen-Yu Lee}, {and}
  \bibinfo{person}{Tomas Pfister}.} \bibinfo{year}{2023}\natexlab{}.
\newblock \showarticletitle{Distilling step-by-step! outperforming larger
  language models with less training data and smaller model sizes}.
\newblock \bibinfo{journal}{\emph{arXiv preprint arXiv:2305.02301}}
  (\bibinfo{year}{2023}).
\newblock


\bibitem[Huang et~al\mbox{.}(2023)]%
        {self_improve}
\bibfield{author}{\bibinfo{person}{Jiaxin Huang},
  \bibinfo{person}{Shixiang~Shane Gu}, \bibinfo{person}{Le Hou},
  \bibinfo{person}{Yuexin Wu}, \bibinfo{person}{Xuezhi Wang},
  \bibinfo{person}{Hongkun Yu}, {and} \bibinfo{person}{Jiawei Han}.}
  \bibinfo{year}{2023}\natexlab{}.
\newblock \bibinfo{title}{Large Language Models Can Self-improve}.
\newblock
\newblock
\urldef\tempurl%
\url{https://openreview.net/forum?id=NiEtU7blzN}
\showURL{%
\tempurl}


\bibitem[Jia et~al\mbox{.}(2024)]%
        {jia2024erase}
\bibfield{author}{\bibinfo{person}{Pengyue Jia}, \bibinfo{person}{Yejing Wang},
  \bibinfo{person}{Zhaocheng Du}, \bibinfo{person}{Xiangyu Zhao},
  \bibinfo{person}{Yichao Wang}, \bibinfo{person}{Bo Chen},
  \bibinfo{person}{Wanyu Wang}, \bibinfo{person}{Huifeng Guo}, {and}
  \bibinfo{person}{Ruiming Tang}.} \bibinfo{year}{2024}\natexlab{}.
\newblock \showarticletitle{Erase: Benchmarking feature selection methods for
  deep recommender systems}. In \bibinfo{booktitle}{\emph{Proceedings of the
  30th ACM SIGKDD Conference on Knowledge Discovery and Data Mining}}.
  \bibinfo{pages}{5194--5205}.
\newblock


\bibitem[Khattab and Zaharia(2020)]%
        {khattab2020colbert}
\bibfield{author}{\bibinfo{person}{Omar Khattab} {and} \bibinfo{person}{Matei
  Zaharia}.} \bibinfo{year}{2020}\natexlab{}.
\newblock \showarticletitle{Colbert: Efficient and effective passage search via
  contextualized late interaction over bert}. In
  \bibinfo{booktitle}{\emph{Proceedings of the 43rd International ACM SIGIR
  conference on research and development in Information Retrieval}}.
  \bibinfo{pages}{39--48}.
\newblock


\bibitem[Leon et~al\mbox{.}(2013)]%
        {Gram_Schmidt}
\bibfield{author}{\bibinfo{person}{Steven~J Leon}, \bibinfo{person}{{\AA}ke
  Bj{\"o}rck}, {and} \bibinfo{person}{Walter Gander}.}
  \bibinfo{year}{2013}\natexlab{}.
\newblock \showarticletitle{Gram-Schmidt orthogonalization: 100 years and
  more}.
\newblock \bibinfo{journal}{\emph{Numerical Linear Algebra with Applications}}
  \bibinfo{volume}{20}, \bibinfo{number}{3} (\bibinfo{year}{2013}),
  \bibinfo{pages}{492--532}.
\newblock


\bibitem[Li et~al\mbox{.}(2023a)]%
        {li2023coannotating}
\bibfield{author}{\bibinfo{person}{Minzhi Li}, \bibinfo{person}{Taiwei Shi},
  \bibinfo{person}{Caleb Ziems}, \bibinfo{person}{Min-Yen Kan},
  \bibinfo{person}{Nancy~F Chen}, \bibinfo{person}{Zhengyuan Liu}, {and}
  \bibinfo{person}{Diyi Yang}.} \bibinfo{year}{2023}\natexlab{a}.
\newblock \showarticletitle{Coannotating: Uncertainty-guided work allocation
  between human and large language models for data annotation}.
\newblock \bibinfo{journal}{\emph{arXiv preprint arXiv:2310.15638}}
  (\bibinfo{year}{2023}).
\newblock


\bibitem[Li et~al\mbox{.}(2023b)]%
        {li2023synthetic}
\bibfield{author}{\bibinfo{person}{Zhuoyan Li}, \bibinfo{person}{Hangxiao Zhu},
  \bibinfo{person}{Zhuoran Lu}, {and} \bibinfo{person}{Ming Yin}.}
  \bibinfo{year}{2023}\natexlab{b}.
\newblock \showarticletitle{Synthetic data generation with large language
  models for text classification: Potential and limitations}.
\newblock \bibinfo{journal}{\emph{arXiv preprint arXiv:2310.07849}}
  (\bibinfo{year}{2023}).
\newblock


\bibitem[Liu et~al\mbox{.}(2024)]%
        {best_practice}
\bibfield{author}{\bibinfo{person}{Ruibo Liu}, \bibinfo{person}{Jerry Wei},
  \bibinfo{person}{Fangyu Liu}, \bibinfo{person}{Chenglei Si},
  \bibinfo{person}{Yanzhe Zhang}, \bibinfo{person}{Jinmeng Rao},
  \bibinfo{person}{Steven Zheng}, \bibinfo{person}{Daiyi Peng},
  \bibinfo{person}{Diyi Yang}, \bibinfo{person}{Denny Zhou}, {et~al\mbox{.}}}
  \bibinfo{year}{2024}\natexlab{}.
\newblock \showarticletitle{Best practices and lessons learned on synthetic
  data}. In \bibinfo{booktitle}{\emph{First Conference on Language Modeling}}.
\newblock


\bibitem[Long et~al\mbox{.}(2024)]%
        {survey_llmsdrivensyntheticdata}
\bibfield{author}{\bibinfo{person}{Lin Long}, \bibinfo{person}{Rui Wang},
  \bibinfo{person}{Ruixuan Xiao}, \bibinfo{person}{Junbo Zhao},
  \bibinfo{person}{Xiao Ding}, \bibinfo{person}{Gang Chen}, {and}
  \bibinfo{person}{Haobo Wang}.} \bibinfo{year}{2024}\natexlab{}.
\newblock \bibinfo{title}{On LLMs-Driven Synthetic Data Generation, Curation,
  and Evaluation: A Survey}.
\newblock
\newblock
\showeprint[arxiv]{2406.15126}~[cs.CL]
\urldef\tempurl%
\url{https://arxiv.org/abs/2406.15126}
\showURL{%
\tempurl}


\bibitem[Lyu et~al\mbox{.}(2022)]%
        {lyu2022semi}
\bibfield{author}{\bibinfo{person}{Yan Lyu}, \bibinfo{person}{Sunhao Dai},
  \bibinfo{person}{Peng Wu}, \bibinfo{person}{Quanyu Dai},
  \bibinfo{person}{Yuhao Deng}, \bibinfo{person}{Wenjie Hu},
  \bibinfo{person}{Zhenhua Dong}, \bibinfo{person}{Jun Xu},
  \bibinfo{person}{Shengyu Zhu}, {and} \bibinfo{person}{Xiao-Hua Zhou}.}
  \bibinfo{year}{2022}\natexlab{}.
\newblock \showarticletitle{A Semi-Synthetic Dataset Generation Framework for
  Causal Inference in Recommender Systems}.
\newblock \bibinfo{journal}{\emph{arXiv preprint arXiv:2202.11351}}
  (\bibinfo{year}{2022}).
\newblock


\bibitem[Meng et~al\mbox{.}(2022)]%
        {fewgen}
\bibfield{author}{\bibinfo{person}{Yu Meng}, \bibinfo{person}{Martin
  Michalski}, \bibinfo{person}{Jiaxin Huang}, \bibinfo{person}{Yu Zhang},
  \bibinfo{person}{Tarek~F. Abdelzaher}, {and} \bibinfo{person}{Jiawei Han}.}
  \bibinfo{year}{2022}\natexlab{}.
\newblock \showarticletitle{Tuning Language Models as Training Data Generators
  for Augmentation-Enhanced Few-Shot Learning}. In
  \bibinfo{booktitle}{\emph{International Conference on Machine Learning}}.
\newblock
\urldef\tempurl%
\url{https://api.semanticscholar.org/CorpusID:253384628}
\showURL{%
\tempurl}


\bibitem[Murphy(2012)]%
        {murphy2012machine}
\bibfield{author}{\bibinfo{person}{Kevin~P Murphy}.}
  \bibinfo{year}{2012}\natexlab{}.
\newblock \bibinfo{booktitle}{\emph{Machine learning: a probabilistic
  perspective}}.
\newblock \bibinfo{publisher}{MIT press}.
\newblock


\bibitem[Nie et~al\mbox{.}(2019)]%
        {RelGAN}
\bibfield{author}{\bibinfo{person}{Weili Nie}, \bibinfo{person}{Nina
  Narodytska}, {and} \bibinfo{person}{Ankit~B. Patel}.}
  \bibinfo{year}{2019}\natexlab{}.
\newblock \showarticletitle{RelGAN: Relational Generative Adversarial Networks
  for Text Generation}. In \bibinfo{booktitle}{\emph{International Conference
  on Learning Representations}}.
\newblock
\urldef\tempurl%
\url{https://api.semanticscholar.org/CorpusID:68160504}
\showURL{%
\tempurl}


\bibitem[Ouyang et~al\mbox{.}(2022)]%
        {ouyang2022training}
\bibfield{author}{\bibinfo{person}{Long Ouyang}, \bibinfo{person}{Jeffrey Wu},
  \bibinfo{person}{Xu Jiang}, \bibinfo{person}{Diogo Almeida},
  \bibinfo{person}{Carroll Wainwright}, \bibinfo{person}{Pamela Mishkin},
  \bibinfo{person}{Chong Zhang}, \bibinfo{person}{Sandhini Agarwal},
  \bibinfo{person}{Katarina Slama}, \bibinfo{person}{Alex Ray},
  {et~al\mbox{.}}} \bibinfo{year}{2022}\natexlab{}.
\newblock \showarticletitle{Training language models to follow instructions
  with human feedback}.
\newblock \bibinfo{journal}{\emph{Advances in neural information processing
  systems}}  \bibinfo{volume}{35} (\bibinfo{year}{2022}),
  \bibinfo{pages}{27730--27744}.
\newblock


\bibitem[Ramasesh et~al\mbox{.}(2021)]%
        {ramasesh2021effect}
\bibfield{author}{\bibinfo{person}{Vinay~Venkatesh Ramasesh},
  \bibinfo{person}{Aitor Lewkowycz}, {and} \bibinfo{person}{Ethan Dyer}.}
  \bibinfo{year}{2021}\natexlab{}.
\newblock \showarticletitle{Effect of scale on catastrophic forgetting in
  neural networks}. In \bibinfo{booktitle}{\emph{International Conference on
  Learning Representations}}.
\newblock


\bibitem[Reimers(2019)]%
        {reimers2019sentence}
\bibfield{author}{\bibinfo{person}{N Reimers}.}
  \bibinfo{year}{2019}\natexlab{}.
\newblock \showarticletitle{Sentence-BERT: Sentence Embeddings using Siamese
  BERT-Networks}.
\newblock \bibinfo{journal}{\emph{arXiv preprint arXiv:1908.10084}}
  (\bibinfo{year}{2019}).
\newblock


\bibitem[Sanh(2019a)]%
        {sanh2019distilbert}
\bibfield{author}{\bibinfo{person}{V Sanh}.} \bibinfo{year}{2019}\natexlab{a}.
\newblock \showarticletitle{DistilBERT, a distilled version of BERT: smaller,
  faster, cheaper and lighter}.
\newblock \bibinfo{journal}{\emph{arXiv preprint arXiv:1910.01108}}
  (\bibinfo{year}{2019}).
\newblock


\bibitem[Sanh(2019b)]%
        {distilbert}
\bibfield{author}{\bibinfo{person}{V Sanh}.} \bibinfo{year}{2019}\natexlab{b}.
\newblock \showarticletitle{DistilBERT, a distilled version of BERT: smaller,
  faster, cheaper and lighter}.
\newblock \bibinfo{journal}{\emph{arXiv preprint arXiv:1910.01108}}
  (\bibinfo{year}{2019}).
\newblock


\bibitem[Seedat et~al\mbox{.}(2023)]%
        {seedat2023curated}
\bibfield{author}{\bibinfo{person}{Nabeel Seedat}, \bibinfo{person}{Nicolas
  Huynh}, \bibinfo{person}{Boris van Breugel}, {and} \bibinfo{person}{Mihaela
  van~der Schaar}.} \bibinfo{year}{2023}\natexlab{}.
\newblock \showarticletitle{Curated llm: Synergy of llms and data curation for
  tabular augmentation in ultra low-data regimes}.
\newblock \bibinfo{journal}{\emph{arXiv preprint arXiv:2312.12112}}
  (\bibinfo{year}{2023}).
\newblock


\bibitem[Shahul~Hameed et~al\mbox{.}(2024)]%
        {shahul2024bias}
\bibfield{author}{\bibinfo{person}{Mohamed~Ashik Shahul~Hameed},
  \bibinfo{person}{Asifa~Mehmood Qureshi}, {and} \bibinfo{person}{Abhishek
  Kaushik}.} \bibinfo{year}{2024}\natexlab{}.
\newblock \showarticletitle{Bias Mitigation via Synthetic Data Generation: A
  Review.}
\newblock \bibinfo{journal}{\emph{Electronics (2079-9292)}}
  \bibinfo{volume}{13}, \bibinfo{number}{19} (\bibinfo{year}{2024}).
\newblock


\bibitem[Socher et~al\mbox{.}(2013)]%
        {sst2}
\bibfield{author}{\bibinfo{person}{Richard Socher}, \bibinfo{person}{Alex
  Perelygin}, \bibinfo{person}{Jean Wu}, \bibinfo{person}{Jason Chuang},
  \bibinfo{person}{Christopher~D Manning}, \bibinfo{person}{Andrew~Y Ng}, {and}
  \bibinfo{person}{Christopher Potts}.} \bibinfo{year}{2013}\natexlab{}.
\newblock \showarticletitle{Recursive deep models for semantic compositionality
  over a sentiment treebank}. In \bibinfo{booktitle}{\emph{Proceedings of the
  2013 conference on empirical methods in natural language processing}}.
  \bibinfo{pages}{1631--1642}.
\newblock


\bibitem[Wan et~al\mbox{.}(2024)]%
        {wan2024tnt}
\bibfield{author}{\bibinfo{person}{Mengting Wan}, \bibinfo{person}{Tara
  Safavi}, \bibinfo{person}{Sujay~Kumar Jauhar}, \bibinfo{person}{Yujin Kim},
  \bibinfo{person}{Scott Counts}, \bibinfo{person}{Jennifer Neville},
  \bibinfo{person}{Siddharth Suri}, \bibinfo{person}{Chirag Shah},
  \bibinfo{person}{Ryen~W White}, \bibinfo{person}{Longqi Yang},
  {et~al\mbox{.}}} \bibinfo{year}{2024}\natexlab{}.
\newblock \showarticletitle{Tnt-llm: Text mining at scale with large language
  models}. In \bibinfo{booktitle}{\emph{Proceedings of the 30th ACM SIGKDD
  Conference on Knowledge Discovery and Data Mining}}.
  \bibinfo{pages}{5836--5847}.
\newblock


\bibitem[Wang and Sennrich(2020)]%
        {wang2020exposure}
\bibfield{author}{\bibinfo{person}{Chaojun Wang} {and} \bibinfo{person}{Rico
  Sennrich}.} \bibinfo{year}{2020}\natexlab{}.
\newblock \showarticletitle{On exposure bias, hallucination and domain shift in
  neural machine translation}.
\newblock \bibinfo{journal}{\emph{arXiv preprint arXiv:2005.03642}}
  (\bibinfo{year}{2020}).
\newblock


\bibitem[Wang et~al\mbox{.}(2023b)]%
        {wang2023let}
\bibfield{author}{\bibinfo{person}{Ruida Wang}, \bibinfo{person}{Wangchunshu
  Zhou}, {and} \bibinfo{person}{Mrinmaya Sachan}.}
  \bibinfo{year}{2023}\natexlab{b}.
\newblock \showarticletitle{Let's Synthesize Step by Step: Iterative Dataset
  Synthesis with Large Language Models by Extrapolating Errors from Small
  Models}.
\newblock \bibinfo{journal}{\emph{arXiv preprint arXiv:2310.13671}}
  (\bibinfo{year}{2023}).
\newblock


\bibitem[Wang et~al\mbox{.}(2023a)]%
        {wang2023single}
\bibfield{author}{\bibinfo{person}{Yejing Wang}, \bibinfo{person}{Zhaocheng
  Du}, \bibinfo{person}{Xiangyu Zhao}, \bibinfo{person}{Bo Chen},
  \bibinfo{person}{Huifeng Guo}, \bibinfo{person}{Ruiming Tang}, {and}
  \bibinfo{person}{Zhenhua Dong}.} \bibinfo{year}{2023}\natexlab{a}.
\newblock \showarticletitle{Single-shot feature selection for multi-task
  recommendations}. In \bibinfo{booktitle}{\emph{Proceedings of the 46th
  International ACM SIGIR Conference on Research and Development in Information
  Retrieval}}. \bibinfo{pages}{341--351}.
\newblock


\bibitem[Wei et~al\mbox{.}(2022)]%
        {wei2022chain}
\bibfield{author}{\bibinfo{person}{Jason Wei}, \bibinfo{person}{Xuezhi Wang},
  \bibinfo{person}{Dale Schuurmans}, \bibinfo{person}{Maarten Bosma},
  \bibinfo{person}{Fei Xia}, \bibinfo{person}{Ed Chi}, \bibinfo{person}{Quoc~V
  Le}, \bibinfo{person}{Denny Zhou}, {et~al\mbox{.}}}
  \bibinfo{year}{2022}\natexlab{}.
\newblock \showarticletitle{Chain-of-thought prompting elicits reasoning in
  large language models}.
\newblock \bibinfo{journal}{\emph{Advances in neural information processing
  systems}}  \bibinfo{volume}{35} (\bibinfo{year}{2022}),
  \bibinfo{pages}{24824--24837}.
\newblock


\bibitem[Williams and Rasmussen(1995)]%
        {williams1995gaussian}
\bibfield{author}{\bibinfo{person}{Christopher Williams} {and}
  \bibinfo{person}{Carl Rasmussen}.} \bibinfo{year}{1995}\natexlab{}.
\newblock \showarticletitle{Gaussian processes for regression}.
\newblock \bibinfo{journal}{\emph{Advances in neural information processing
  systems}}  \bibinfo{volume}{8} (\bibinfo{year}{1995}).
\newblock


\bibitem[Wu et~al\mbox{.}(2008)]%
        {wu2008information}
\bibfield{author}{\bibinfo{person}{Fei Wu}, \bibinfo{person}{Raphael Hoffmann},
  {and} \bibinfo{person}{Daniel~S Weld}.} \bibinfo{year}{2008}\natexlab{}.
\newblock \showarticletitle{Information extraction from Wikipedia: Moving down
  the long tail}. In \bibinfo{booktitle}{\emph{Proceedings of the 14th ACM
  SIGKDD international conference on Knowledge discovery and data mining}}.
  \bibinfo{pages}{731--739}.
\newblock


\bibitem[Xiao et~al\mbox{.}(2023)]%
        {xiao2023freeal}
\bibfield{author}{\bibinfo{person}{Ruixuan Xiao}, \bibinfo{person}{Yiwen Dong},
  \bibinfo{person}{Junbo Zhao}, \bibinfo{person}{Runze Wu},
  \bibinfo{person}{Minmin Lin}, \bibinfo{person}{Gang Chen}, {and}
  \bibinfo{person}{Haobo Wang}.} \bibinfo{year}{2023}\natexlab{}.
\newblock \showarticletitle{Freeal: Towards human-free active learning in the
  era of large language models}.
\newblock \bibinfo{journal}{\emph{arXiv preprint arXiv:2311.15614}}
  (\bibinfo{year}{2023}).
\newblock


\bibitem[Ye et~al\mbox{.}(2022)]%
        {progen}
\bibfield{author}{\bibinfo{person}{Jiacheng Ye}, \bibinfo{person}{Jiahui Gao},
  \bibinfo{person}{Jiangtao Feng}, \bibinfo{person}{Zhiyong Wu},
  \bibinfo{person}{Tao Yu}, {and} \bibinfo{person}{Lingpeng Kong}.}
  \bibinfo{year}{2022}\natexlab{}.
\newblock \showarticletitle{Progen: Progressive zero-shot dataset generation
  via in-context feedback}.
\newblock \bibinfo{journal}{\emph{arXiv preprint arXiv:2210.12329}}
  (\bibinfo{year}{2022}).
\newblock


\bibitem[Yoo et~al\mbox{.}(2021)]%
        {GPT3Mix}
\bibfield{author}{\bibinfo{person}{Kang~Min Yoo}, \bibinfo{person}{Dongju
  Park}, \bibinfo{person}{Jaewook Kang}, \bibinfo{person}{Sang-Woo Lee}, {and}
  \bibinfo{person}{Woomyeong Park}.} \bibinfo{year}{2021}\natexlab{}.
\newblock \showarticletitle{GPT3Mix: Leveraging Large-scale Language Models for
  Text Augmentation}. In \bibinfo{booktitle}{\emph{Conference on Empirical
  Methods in Natural Language Processing}}.
\newblock
\urldef\tempurl%
\url{https://api.semanticscholar.org/CorpusID:233296100}
\showURL{%
\tempurl}


\bibitem[Yu et~al\mbox{.}(2023a)]%
        {attriprompt}
\bibfield{author}{\bibinfo{person}{Yue Yu}, \bibinfo{person}{Yuchen Zhuang},
  \bibinfo{person}{Jieyu Zhang}, \bibinfo{person}{Yu Meng},
  \bibinfo{person}{Alexander~J. Ratner}, \bibinfo{person}{Ranjay Krishna},
  \bibinfo{person}{Jiaming Shen}, {and} \bibinfo{person}{Chao Zhang}.}
  \bibinfo{year}{2023}\natexlab{a}.
\newblock \showarticletitle{Large Language Model as Attributed Training Data
  Generator: A Tale of Diversity and Bias}.
\newblock \bibinfo{journal}{\emph{ArXiv}}  \bibinfo{volume}{abs/2306.15895}
  (\bibinfo{year}{2023}).
\newblock
\urldef\tempurl%
\url{https://api.semanticscholar.org/CorpusID:259275123}
\showURL{%
\tempurl}


\bibitem[Yu et~al\mbox{.}(2023b)]%
        {regen}
\bibfield{author}{\bibinfo{person}{Yue Yu}, \bibinfo{person}{Yuchen Zhuang},
  \bibinfo{person}{Rongzhi Zhang}, \bibinfo{person}{Yu Meng},
  \bibinfo{person}{Jiaming Shen}, {and} \bibinfo{person}{Chao Zhang}.}
  \bibinfo{year}{2023}\natexlab{b}.
\newblock \showarticletitle{Regen: Zero-shot text classification via training
  data generation with progressive dense retrieval}.
\newblock \bibinfo{journal}{\emph{arXiv preprint arXiv:2305.10703}}
  (\bibinfo{year}{2023}).
\newblock


\bibitem[Zhang et~al\mbox{.}(2015)]%
        {agnews}
\bibfield{author}{\bibinfo{person}{Xiang Zhang}, \bibinfo{person}{Junbo Zhao},
  {and} \bibinfo{person}{Yann LeCun}.} \bibinfo{year}{2015}\natexlab{}.
\newblock \showarticletitle{Character-level convolutional networks for text
  classification}.
\newblock \bibinfo{journal}{\emph{NeurIPS}}  \bibinfo{volume}{28}
  (\bibinfo{year}{2015}).
\newblock


\bibitem[Zhao and Bilen(2023)]%
        {zhao2023dataset}
\bibfield{author}{\bibinfo{person}{Bo Zhao} {and} \bibinfo{person}{Hakan
  Bilen}.} \bibinfo{year}{2023}\natexlab{}.
\newblock \showarticletitle{Dataset condensation with distribution matching}.
  In \bibinfo{booktitle}{\emph{Proceedings of the IEEE/CVF Winter Conference on
  Applications of Computer Vision}}. \bibinfo{pages}{6514--6523}.
\newblock


\bibitem[Zhao et~al\mbox{.}(2024)]%
        {zhao2024retrievable}
\bibfield{author}{\bibinfo{person}{Yuang Zhao}, \bibinfo{person}{Zhaocheng Du},
  \bibinfo{person}{Qinglin Jia}, \bibinfo{person}{Linxuan Zhang},
  \bibinfo{person}{Zhenhua Dong}, {and} \bibinfo{person}{Ruiming Tang}.}
  \bibinfo{year}{2024}\natexlab{}.
\newblock \showarticletitle{Retrievable Domain-Sensitive Feature Memory for
  Multi-Domain Recommendation}.
\newblock \bibinfo{journal}{\emph{arXiv preprint arXiv:2405.12892}}
  (\bibinfo{year}{2024}).
\newblock


\end{thebibliography}
